\documentclass[10pt,twocolumn,letterpaper]{article}

\usepackage[pagenumbers]{iccv} 

\usepackage{graphicx}	
\usepackage{amsmath}	
\usepackage{amssymb}	
\usepackage{booktabs}
\usepackage{cuted}
\usepackage{times}
\usepackage{microtype}
\usepackage{epsfig}
\usepackage{caption}
\usepackage{float}
\usepackage{placeins}
\usepackage{color, colortbl}
\usepackage{stfloats}
\usepackage{enumitem}
\usepackage{tabularx}
\usepackage{xstring}
\usepackage{multirow}
\usepackage{xspace}
\usepackage{url}
\usepackage{subcaption}
\usepackage[hang,flushmargin]{footmisc}
\usepackage{kotex}
\usepackage{arydshln}
\usepackage{adjustbox}
\usepackage{wrapfig}
\usepackage{algorithm,algorithmicx,algpseudocode}
\usepackage{listings}
\usepackage{bm}

\newlength\paramargin
\newlength\abovetabcapmargin
\newlength\belowtabcapmargin
\newlength\abovefigcapmargin
\newlength\belowfigcapmargin
\newlength\aboveeqmargin
\newlength\beloweqmargin

\definecolor{iccvblue}{rgb}{0.21,0.49,0.74}
\usepackage[pagebackref,breaklinks,colorlinks,allcolors=iccvblue]{hyperref}

\usepackage{kotex}

\def\paperID{7919} 
\def\confName{ICCV}
\def\confYear{2025}

\title{VideoRFSplat: Direct Scene-Level Text-to-3D Gaussian Splatting Generation with Flexible Pose and Multi-View Joint Modeling}

\author{
    Hyojun Go\textsuperscript{\rm1}\thanks{Equal contribution, $^\dagger$Corresponding author}  \qquad
    Byeongjun Park\textsuperscript{\rm1,\rm2}\footnotemark[1] \qquad
    Hyelin Nam\textsuperscript{\rm1} \qquad
    Byung-Hoon Kim\textsuperscript{\rm1,\rm3} \\ 
    Hyungjin Chung\textsuperscript{\rm1$\dagger$} \qquad
    Changick Kim\textsuperscript{\rm2$\dagger$} \vspace{0.45em} \\
    \textsuperscript{\rm 1} EverEx \qquad
    \textsuperscript{\rm 2} KAIST \qquad
    \textsuperscript{\rm 3} Yonsei University
    \vspace{0.45em} \\
    \href{https://gohyojun15.github.io/VideoRFSplat/}{https://gohyojun15.github.io/VideoRFSplat/}    
}

\begin{document}

\maketitle

    
    

\begin{strip}
    \centering
    \vspace{-5.55em}
    \resizebox{\textwidth}{!}{
    \includegraphics[width=\textwidth]{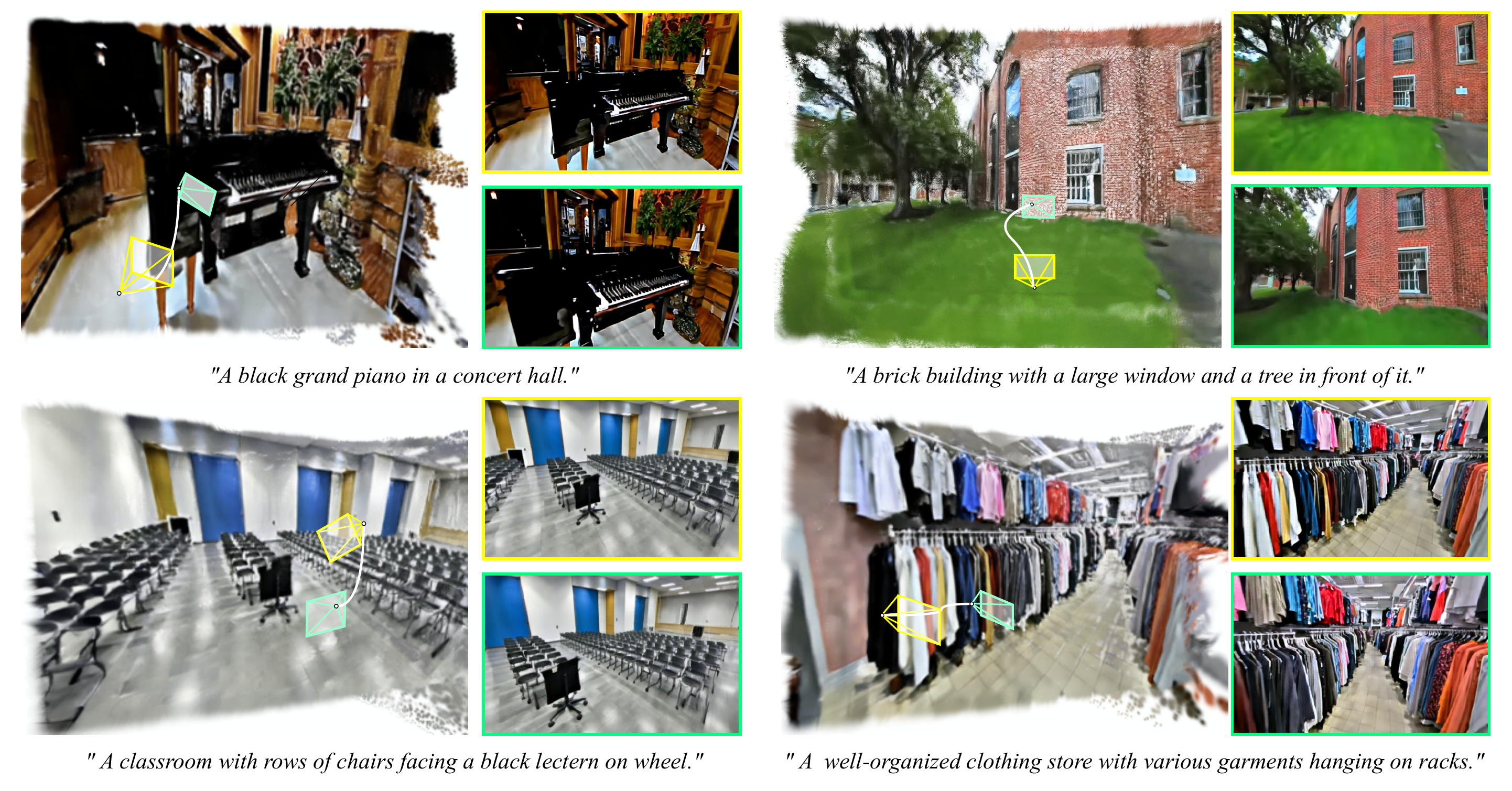}  
    }
    \vspace{\abovefigcapmargin}
    \vspace{-0.55cm}
    \captionof{figure}{\textbf{Generated 3D Gaussian Splattings and rendered views from diverse texts by VideoRFSplat.} VideoRFSplat directly generates realistic 3D scenes from text without SDS~\cite{poole2023dreamfusion, li2025director3d} refinement, outperforming prior methods~\cite{go2024splatflow, li2025director3d} that rely on SDS refinements.}
    \label{fig:enter-label}
    \vspace{-0.15cm}
    \vspace{\abovefigcapmargin}
\end{strip}

\begin{abstract}
We propose VideoRFSplat, a direct text-to-3D model leveraging a video generation model to generate realistic 3D Gaussian Splatting (3DGS) for unbounded real-world scenes.
To generate diverse camera poses and unbounded spatial extent of real-world scenes, while ensuring generalization to arbitrary text prompts, previous methods fine-tune 2D generative models to jointly model camera poses and multi-view images.
However, these methods suffer from instability when extending 2D generative models to joint modeling due to the modality gap, which necessitates additional models to stabilize training and inference.
In this work, we propose an architecture and a sampling strategy to jointly model multi-view images and camera poses when fine-tuning a video generation model.
Our core idea is a dual-stream architecture that attaches a dedicated pose generation model alongside a pre-trained video generation model via communication blocks, generating multi-view images and camera poses through separate streams.
This design reduces interference between the pose and image modalities.
Additionally, we propose an asynchronous sampling strategy that denoises camera poses faster than multi-view images, allowing rapidly denoised poses to condition multi-view generation, reducing mutual ambiguity and enhancing cross-modal consistency.
Trained on multiple large-scale real-world datasets (RealEstate10K, MVImgNet, DL3DV-10K, ACID), VideoRFSplat outperforms existing text-to-3D direct generation methods that heavily depend on post-hoc refinement via score distillation sampling, achieving superior results without such refinement.
\vspace{-0.5cm}
\end{abstract}

\section{Introduction}
Generating realistic 3D scenes from text has garnered increasing attention in AR, gaming, and robotics. 
Early works~\cite{zhang2024text2nerf, lin2023magic3d} primarily relied on Neural Radiance Fields (NeRF)~\cite{mildenhall2021nerf, muller2022instant, poole2023dreamfusion} to model 3D scenes, but their computationally intensive volumetric rendering poses limitations for real-time rendering. 
Recently, 3D Gaussian Splatting (3DGS)~\cite{kerbl20233d} has emerged as a promising alternative, enabling real-time rendering while preserving high-fidelity details. 
Consequently, recent 3D generation approaches~\cite{chen2024text, li2024connecting,  Yi_2024_CVPR, Ling_2024_CVPR, tang2024lgm, xu2024grm, zhang2024gs} have increasingly adopted 3DGS for its efficiency and high-quality results.

To generate 3DGS, several approaches~\cite{chen2024text, li2024connecting, liu2024humangaussian, tang2024dreamgaussian, vilesov2023cg3d, Yi_2024_CVPR, Ling_2024_CVPR} use Score Distillation Sampling (SDS)~\cite{poole2023dreamfusion} with 2D diffusion models~\cite{Rombach_2022_CVPR}, but this requires time-expensive per-scene optimization. 
Another line of work adopts pipeline-based methods, where generated images are reconstructed into 3DGS using separate reconstruction models~\cite{tang2024lgm, xu2024grm, zhang2024gs} or multi-stage frameworks~\cite{yu2024wonderworld, chung2023luciddreamer, shriram2024realmdreamer, yu2024wonderjourney, zhou2024dreamscene360}.
While pipeline-based methods can produce high-quality results, their reliance on sequential processing increases the risk of failure and adds unnecessary complexity~\cite{lin2025diffsplat}.
To overcome these limitations, recent works have explored direct 3DGS generation~\cite{lin2025diffsplat, he2024gvgen, zhang2024gaussiancube, go2024splatflow, li2025director3d, yang2024prometheus}, which enables end-to-end inference, eliminating per-scene optimization and complex multi-stage pipelines.
A key approach in this direction is fine-tuning 2D generative models~\cite{Rombach_2022_CVPR, genmo2024mochi, flux2024, kong2024hunyuanvideo}, which possess strong text-to-2D generation capabilities, to extend their generalizability to 3DGS from arbitrary texts.

Despite the growing number of works targeting 3D generation, the majority focuses on object-level synthesis, leaving real-world scene generation comparatively underexplored. 
Object-level generation typically operates in controlled, predefined settings as seen in  Objaverse datasets~\cite{deitke2023objaverse, deitke2023objaversexl}, where objects are captured with fixed or simple camera trajectories.
In contrast, real-world scenes present additional challenges due to their complex and diverse scene-specific camera motions, as well as their unbounded nature with diverse backgrounds and large-scale spatial variations~\cite{li2025director3d, go2024splatflow}.
These pose fundamental challenges to developing generative models for direct 3DGS generation, introducing difficulties distinct from object-level generation.

To address these challenges, previous works~\cite{go2024splatflow, li2025director3d} have adopted generative models that learn the joint distribution of multi-view images and camera trajectories.
By incorporating the camera pose into the generative formulation, the model (1) implicitly infers the scene’s pose from the text prompt, allowing it to handle diverse and complex camera poses without requiring user inputs. 
(2) The model generates consistent multi-view images aligned with these inferred scene-specific viewpoints, effectively capturing unbounded environments with diverse backgrounds and spatial variations. 


However, prior works~\cite{go2024splatflow, li2024nvcomposer,li2025director3d} have suffered from instability in extending 2D generative models to joint modeling due to the modality gap, hindering high-quality generation and alignment between generated images and camera poses.
To address this, several stabilization techniques leveraging external models have been proposed.
For instance, NVComposer~\cite{li2024nvcomposer} leverages Dust3R~\cite{wang2024dust3r} distillation to improve consistency, 
while SplatFlow~\cite{go2024splatflow} and director3D~\cite{li2025director3d} rely on other 2D models and refinements during sampling.
While these help reduce instability, dependency on external models hinders seamless integration into a single model.

In this paper, to eliminate external dependency, we present \textit{VideoRFSplat}, a direct 3DGS generation model that introduces an architecture and sampling strategy for jointly generating camera poses and multi-view images when leveraging a video generation model.
Our core idea is a dual-stream architecture, side-attaching a dedicated rectified flow-based pose generation model alongside a pre-trained video generation model, jointly trained to generate multi-view image latents and camera poses simultaneously.
This side-attached pose diffusion model runs parallel to the video model’s forward stream, interacting at specific layers while maintaining separate forward paths. 
This design minimizes interference between the two modalities, allowing each to specialize independently and ensuring consistency between poses and multi-view images.
Similar to previous approaches~\cite{yang2024prometheus, go2024splatflow, li2025director3d}, a Gaussian splat decoder decodes 3DGS from the generated poses and image latents in a feed-forward fashion.

Then, we decouple the timesteps of the pose and multi-view generation models, allowing each to operate at different noise levels independently.
Unlike standard approaches that synchronize timesteps and noise levels across modalities, our design permits flexible asynchronous sampling. 
This approach is motivated by our observation that synchronized denoising of multi-view images and camera poses, particularly at early timesteps, leads to mutual ambiguity, increasing uncertainty and causing unstable generation.
To mitigate this issue, we design the pose modality—found to be more robust to faster denoising—to undergo a more rapid denoising process than the images.
By doing so, the clearer pose information effectively reduces the ambiguity in the pose modality, stabilizing the sampling.
Furthermore, we propose an asynchronous adaptation of Classifier-Free Guidance (CFG) that enables the clearer pose to better guide multi-view image generation.
Moreover, the proposed asynchronous sampling strategy with decoupled timesteps naturally extends to the camera conditional generation task.



We train VideoRFSplat on RealEstate10K~\cite{zhou2018stereo}, MVImgNet~\cite{yu2023mvimgnet}, DL3DV-10K~\cite{ling2024dl3dv}, and ACID~\cite{liu2021infinite} datasets.
Notably, VideoRFSplat achieves superior performance without relying on SDS++ refinement~\cite{li2025director3d}, surpassing existing text-to-3D direct generation methods that depend on SDS++~\cite{li2025director3d, go2024splatflow}, demonstrating effectiveness of our approach and eliminating dependencies on external models.

\section{Related Works}

\subsection{Diffusion Models and Rectified Flow}

Diffusion models~\cite{ho2020denoising, sohl2015deep,song2020score} have recently achieved state-of-the-art performance in high-fidelity generation across diverse modalities, including images~\cite{dhariwal2021diffusion, go2023addressing, park2024switch, park2024denoising, lee2024multi}, videos~\cite{ho2022video, singer2022make,blattmann2023align}, and text~\cite{nie2025large}.
In particular, large-scale text-to-video diffusion models~\cite{yang2024cogvideox, blattmann2023stable, hong2022cogvideo} have gained attention, trained on extensive text-video datasets to support arbitrary text inputs. 
These models inherently possess the capability to generate multiple coherent frames, making them attractive for adaptation to 3D tasks such as image-to-3D generation~\cite{yu2024viewcrafter, liu2024reconx, liang2024wonderland, lin2025diffsplat}.

Rectified Flow (RF) models~\cite{liu2022flow, albergo2022building, lipman2022flow} provide an alternative to conventional diffusion methods by defining a linear transition from data to noise. 
Motivated by the success of Stable Diffusion 3~\cite{esser2024scaling} in scaling RF for text-to-image generation, recent generative models for images and videos~\cite{genmo2024mochi, flux2024, kong2024hunyuanvideo} have increasingly adopted RF. 
Following this, we leverage Mochi~\cite{genmo2024mochi}, a high-performing RF-based video generative model, as the backbone for our method.

\subsection{3D Generative Models}
\paragraph{Scene Representation.}
Early 3D generation methods synthesize explicit representations such as point clouds~\cite{nichol2022point, mo2023dit, vahdat2022lion}, shapes~\cite{Zhou_2021_ICCV}, and voxel grids~\cite{sanghi2023clip, ren2024xcube}. 
However, these approaches typically struggle to achieve photorealistic outcomes. 
Neural Radiance Fields (NeRF)~\cite{mildenhall2021nerf, muller2022instant} significantly advanced photorealism in 3D generation~\cite{chen2023single, poole2023dreamfusion}, but their computationally intensive volumetric rendering limits real-time applications. 
Recently, 3D Gaussian Splatting (3DGS)~\cite{kerbl20233d} has emerged as a promising alternative, enabling real-time rendering and high visual fidelity. 
Building on these advances, our method utilizes 3DGS to achieve rendering efficiency and visual quality.

\vspace{\paramargin}
\paragraph{Text-to-3DGS Generation.} 
Several methods~\cite{chen2024text, tang2024dreamgaussian} leverage Score Distillation Sampling (SDS)~\cite{poole2023dreamfusion} for 3DGS generation, requiring expensive per-scene optimization.
Alternatively, pipeline-based methods first generate images and then reconstruct them into 3DGS representations using separate 3DGS reconstruction models~\cite{tang2024lgm, xu2024grm, zhang2024gs}.
While these methods accelerate generation compared to SDS-based methods, their sequential processes introduce complexity and susceptibility to failure~\cite{lin2025diffsplat}.
To further simplify and improve stability, recent works have shifted towards direct 3DGS generation models, enabling amortized, end-to-end inference~\cite{lin2025diffsplat, he2024gvgen, zhang2024gaussiancube, go2024splatflow, li2025director3d, yang2024prometheus}. 
However, the majority of these direct generation methods~\cite{lin2025diffsplat,zhang2024gaussiancube,he2024gvgen} focus on object-level synthesis, typically utilizing datasets with consistent, canonical viewpoints such as Objaverse datasets~\cite{deitke2023objaverse, deitke2023objaversexl}.

In contrast, real-world scene generation presents greater challenges due to variations in scale, complexity, and scene-specific camera trajectories.
Early methods for real-world scene generation~\cite{zhou2024dreamscene360, yu2024wonderworld, chung2023luciddreamer, shriram2024realmdreamer, yu2024wonderjourney} employ progressive depth-based warping and diffusion-based inpainting, suffering from distortions due to per-view inpainting. 
Another approach conditions models on predefined camera poses~\cite{yang2024prometheus} or both images and poses~\cite{liang2024wonderland}, but relies on manual or heuristic camera trajectories, limiting flexibility for diverse scene-specific motions.
Alternatively, joint modeling methods have recently been proposed to simultaneously generate multi-view images and corresponding camera trajectories directly from text prompts~\cite{li2025director3d, go2024splatflow}. 
For instance, Director3D~\cite{li2025director3d} first generates camera trajectories and then synthesizes images, while SplatFlow~\cite{go2024splatflow} generates both concurrently within a unified model. 
However, to achieve high-quality generation, these methods rely on SDS refinement~\cite{li2025director3d} or incorporate additional models during sampling, hindering seamless integration into a single model.
In this work, we propose a sampling and architectural approach that enables superior 3DGS generation without such refinements.





\section{Preliminary: Rectified Flow}

Rectified Flows (RFs)~\cite{liu2022flow, albergo2022building, lipman2022flow} define a generative model that starts from a simple noise distribution \(q_0\) (often $\mathcal{N}(0, I)$) and evolves samples via an ODE:
\vspace{\aboveeqmargin}
\vspace{-0.1cm}
\begin{equation}
    \mathrm{d}X_t = v_t(X_t)\,\mathrm{d}t,\quad X_0 \sim q_0,\quad t \in [0,1],
\vspace{\beloweqmargin}
\vspace{-0.1cm}
\end{equation}
where $v_t(X_t)$ is modeled by a neural network $v_t(X_t) = -\,u_\theta(X_t,\,1-t)$. To train $u_\theta$, one pairs samples from the data distribution $p_0$ and noise distribution $q_0$ (collectively denoted $p_1$) and considers a linear path $Y_t = tY_1 + (1-t)Y_0$. This induces the ODE $\mathrm{d}Y_t = u_t(Y_t\!\mid\!Y_1)\,\mathrm{d}t$ with $u_t(Y_t\!\mid\!Y_1)=Y_1 - Y_0$, and the marginal vector field is
\vspace{\aboveeqmargin}
\vspace{-0.1cm}
\begin{equation}
\label{eq:margin_vector}
    u_t(Y_t) 
    = \int \!u_t({Y}_t\!\mid\!Y_1)\,\frac{p_t(Y_t\!\mid\!Y_1)}{p_t(Y_t)}\,p_1(Y_1)\,\mathrm{d}Y_1.
\vspace{\beloweqmargin}
\vspace{-0.1cm}
\end{equation}
Minimizing the flow-matching loss $\mathcal{L}_{\text{FM}} = \mathbb{E}_{t,Y_t}\!\bigl[\|u_t(Y_t) - u_\theta(Y_t,t)\|^2\bigr]$
is typically replaced by the more efficient conditional flow-matching objective:
\vspace{\aboveeqmargin}
\vspace{-0.1cm}
\begin{equation}
    \mathcal{L}_{\text{CFM}} 
    = \mathbb{E}_{t,Y_1\sim p_1}\!\bigl[\|u_t(Y_t\!\mid\!Y_1) - u_\theta(Y_t,t)\|^2\bigr],
\vspace{-0.1cm}
\vspace{\beloweqmargin}
\end{equation}
whose gradients coincide with those of $\mathcal{L}_{\text{FM}}$~\cite{lipman2022flow}. After training, samples are generated by solving the ODE $\mathrm{d}X_t = -\,u_\theta(X_t,\,1-t)\,\mathrm{d}t$ from $t=0$ to $t=1$.

\begin{figure*}
    \centering
    \includegraphics[width=0.88\textwidth]{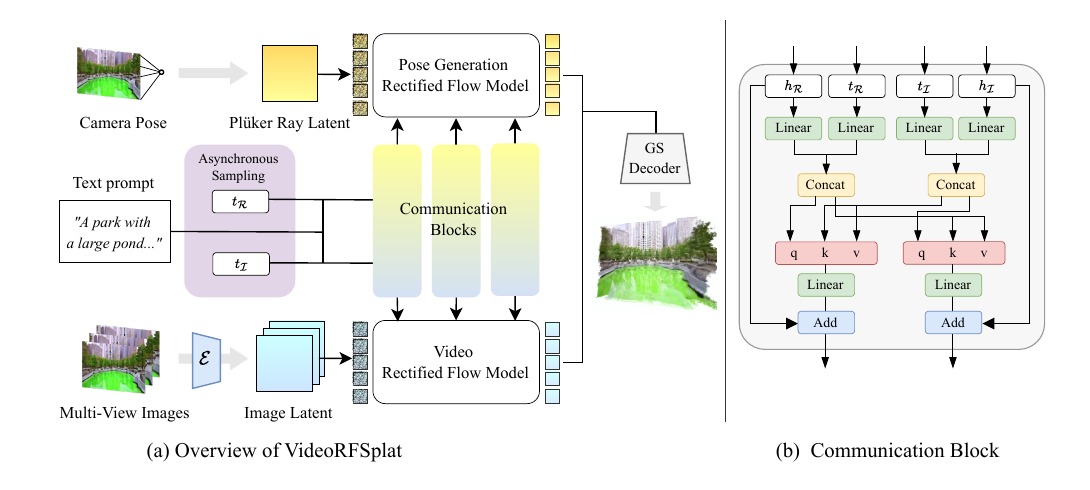}
    \vspace{-0.5cm}
    \caption{
    \textbf{VideoRFSplat Overview.}
    (a) VideoRFSplat consists of a dual-stream pose-video model and a Gaussian Splat decoder.
    To minimize pose-image interference, the pose model is side-attached to the pre-trained video model, interacting through communication blocks.
    With separate timesteps for pose and video models, this enables asynchronous sampling, reducing ambiguity and improving sampling stability.
    (b) Communication block, where cross-attention facilitates bidirectional information exchange between the pose and image modalities.
    }
    \vspace{-0.4cm}
    \label{fig:method}
\end{figure*}
\section{VideoRFSplat}

In this section, we present VideoRFSplat, a framework for generating 3DGS from text prompts for real-world scenes.
As illustrated in Fig.~\ref{fig:method}, our model consists of two main components:
1) A dual-stream pose-video joint model, where a pose generation module is side-attached to a pretrained video generation model, jointly trained to generate multi-view image latents and camera poses.
2) A Gaussian Splat decoder, which reconstructs 3DGS from the generated image latents and camera poses in a feed-forward manner.
In Section~\ref{sec:mvrf_model}, we first present the architecture and sampling process of the dual-stream pose-video joint model. 
Then, Section~\ref{sec:GSdecoder} details the Gaussian Splat decoder.

\subsection{Dual-Stream Pose-Video Joint Model}
\label{sec:mvrf_model}

\paragraph{Architecture.}
We start by considering a set of $K$-view images $\mathcal{I}^o = \{\bm{I}_i\}_{i=1}^{K}$ ($\bm{I}_i \in \mathbb{R}^{H\times W\times 3}$) and their camera projection matrices $\mathcal{P} = \{\bm{P}_i\}_{i=1}^{K} (\bm{P}_i = \mathbf{K}_i[\mathbf{R}_i|\mathbf{T}_i]$), where $\mathbf{K}_i$, $\mathbf{R}_i$, and $\mathbf{T}_i$ denote 
intrinsic, rotation, translation parameters.
Then, we define the latents of the images as $\mathcal{I}=\mathcal{E} (I_0,..., I_K)$.
The goal of the dual-stream pose-video joint model is to generate $\mathcal{I}$ and $\mathcal{P}$ simultaneously.
Building on RayDiffusion~\cite{zhang2024cameras}, which demonstrated the effectiveness of modeling camera poses as rays, and on recent joint-generation methods~\cite{go2024splatflow, li2024nvcomposer}, we adopt the Plücker ray representation~\cite{plucker1828analytisch, zhang2024cameras}.
Each ray $\bm{r}_i$ is parameterized as $\langle \bm{d}_i, \bm{m}_i \rangle$, where $\bm{d}_i=\mathbf{R}_i^\top \mathbf{K}_i^{-1} \bm{w}_i$ and $\bm{m}_i = (-\mathbf{R}_i^\top \mathbf{T}_i) \times \bm{d}_i$.
Here, $\bm{w}_i$ denotes the corresponding 2D pixel coordinates.
For brevity, we denote the full set of rays as $\mathcal{R}=\{\bm{r}_i\}_{i=1}^{K}$.

To extend pre-trained 2D diffusion models for joint generation of camera poses and multi-view images, previous works~\cite{go2024splatflow, li2024nvcomposer, yang2024prometheus} typically concatenate image latents with camera ray embeddings by aligning their spatial resolutions. 
However, we observe that this approach degrades text generalization and overall generation quality (see Table~\ref{tab:architecture_comparison} and Fig.~\ref{fig:architectural_comparison}).  
We attribute this degradation to interference from sharing the same forward path and parameters for image and camera pose. 
Specifically, since pre-trained video models are trained on natural video distributions, incorporating Plücker ray embeddings within this shared structure introduces conflicts due to the significant distribution gap, further degrading text generalization and generation quality.
As a result, this architectural limitation prevents fully leveraging the capabilities of 2D models, leading SplatFlow~\cite{go2024splatflow} to rely on SDS for improved quality.

To reduce interference, we propose a dual-stream architecture with dedicated submodules for pose and image generation, communicating via cross-attention at intermediate layers (see Fig.~\ref{fig:method} (a)).
The pose generation model adopts a transformer-based architecture~\cite{vaswani2017attention,genmo2024mochi}, explicitly conditioned on textual prompts and pose-specific timestep to generate camera rays~\cite{zhang2024cameras}, forming a structurally self-contained module for camera pose generation.
Communication between the two modules occurs at specific intermediate layers through cross-attention mechanisms (see Fig.~\ref{fig:method} (b)). 
Specifically, let $h_\mathcal{R}$ and $h_\mathcal{I}$ be the intermediate outputs of the pose and video models, respectively.
We concatenate the pose timestep embedding and $h_\mathcal{R}$, as well as the video timestep embedding and $h_\mathcal{I}$, then update them bidirectionally via cross-attention: $h_\mathcal{R} \leftarrow \text{CrossAttention}(h_\mathcal{I}, h_\mathcal{R})$ and $h_{\mathcal{I}} \leftarrow \text{CrossAttention}(h_\mathcal{R}, h_\mathcal{I})$.
This exchange enables controlled interaction between the two models while preserving their specialized forward paths and reducing interference between pose and multi-view modalities.
Further details on the architecture are provided in Appendix~\textcolor{iccvblue}{A}.

\begin{figure}
    \centering
    \includegraphics[width=\linewidth]{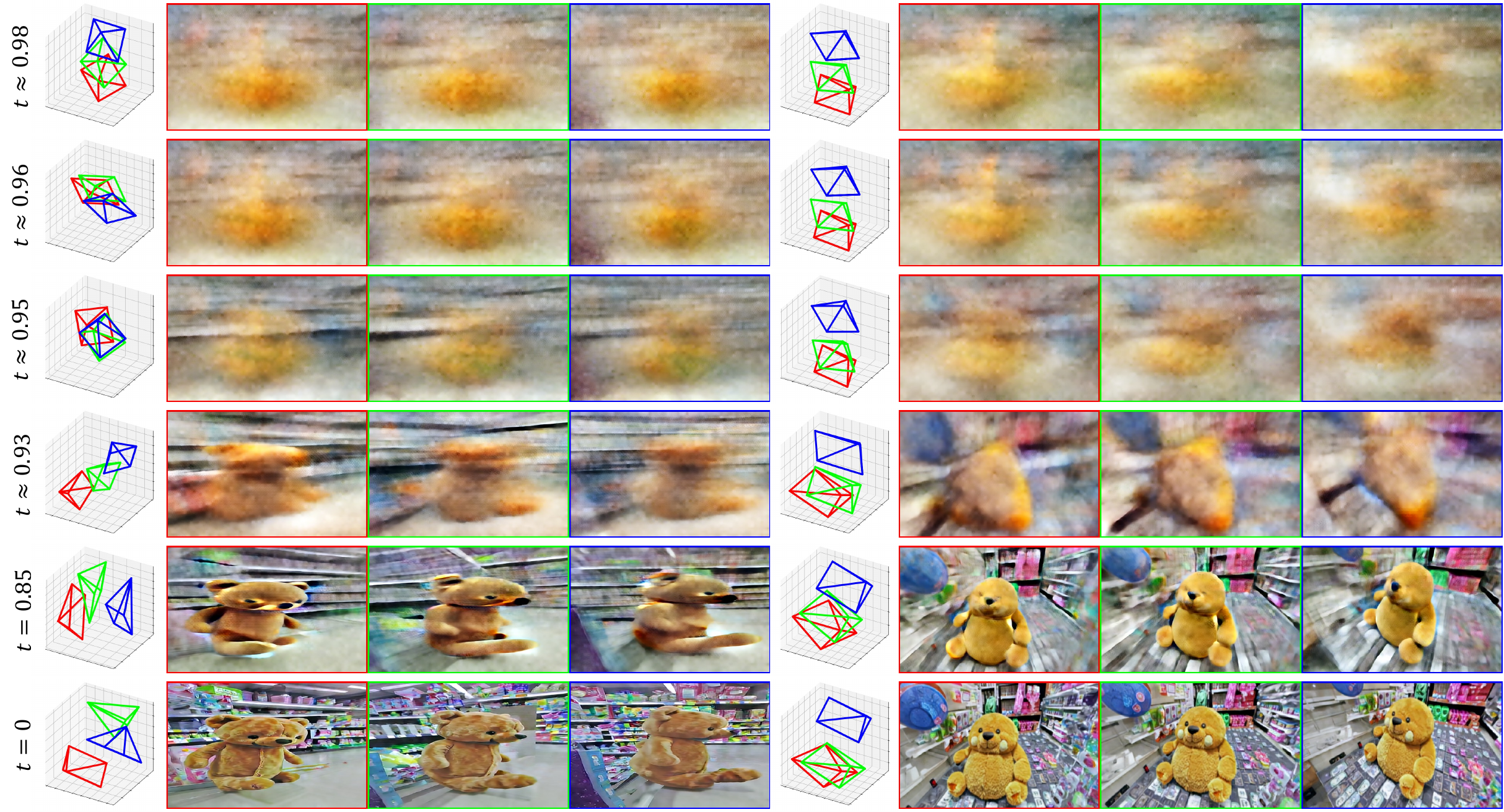}
    \vspace{-0.7cm}
    \caption{\textbf{Failure analysis of synchronized sampling and the effectiveness of asynchronous sampling.}
    \textbf{\textit{(Left)} }
    Early in sampling ($t > 0.85$), 
    synchronous sampling induces excessive oscillations in camera poses, causing divergence and misalignment with images.
    Then, misaligned poses lead to inconsistent multi-view generation, particularly in the background. \textbf{\textit{(Right)}} Asynchronous sampling stabilizes joint generation, leading to coherent multi-view generation.
    }
    \vspace{-0.1cm}
    \label{fig:unstability}
\end{figure}







\vspace{-0.05cm}
\vspace{\paramargin}
\paragraph{Asynchronous Sampling.}
Following Eq.~\ref{eq:margin_vector}, joint modeling of multi-view image latents $\mathcal{I}$ and their corresponding camera rays $\mathcal{R}$ involves estimating the conditional expectation of the marginal vector field at timestep $t$, represented as $\mathbb{E}[u_t^{\mathcal{I}}, u_t^{\mathcal{R}}\mid \mathcal{I}_t,\mathcal{R}_t]$, where $u_t^{\mathcal{I}}$ and $u_t^{\mathcal{R}}$ are conditional vector fields for multi-view images and camera rays conditioned on text prompts.
Here, $\mathcal{I}_t$ and $\mathcal{R}_t$ represent the corresponding intermediate noisy data at timestep $t$.  
Previous joint modeling approaches~\cite{go2024splatflow, li2024nvcomposer} adopt a synchronous timestep strategy, assuming a same timestep for both modalities.  
However, during joint sampling, images and camera poses are inferred together, creating a mutual dependency.  
In early sampling, both modalities are noisy and incomplete, each relying on ambiguous signals from the other.
This mutual ambiguity amplifies uncertainty, potentially causing divergence and unstable generation outcomes, as illustrated in Fig.~\ref{fig:unstability}.

To address this, we propose an asynchronous timestep strategy, decoupling the timesteps of pose and multi-view generation modules and enabling one modality to denoise faster, thereby reducing ambiguity during sampling.
To enable this, we use the following loss:
\vspace{\aboveeqmargin}
\vspace{-0.1cm}
\begin{equation}
~\label{eq:timestep_loss}
    \mathcal{L}_{ours} := \mathbb{E}_{t_{\mathcal{I}}, t_{\mathcal{R}}}\left[ \| [u_{t_{\mathcal{I}}}^{\mathcal{I}}, u_{t_{\mathcal{R}}}^{\mathcal{R}}] - u_\theta(\mathcal{I}_{t_{\mathcal{I}}}, \mathcal{R}_{t_{\mathcal{R}}}, c, t_{\mathcal{I}}, t_{\mathcal{R}})\|^2 \right],
\vspace{\beloweqmargin}
\end{equation}
where $\theta$ denotes parameters, $c$ is the conditioning text prompt, and we omit the expectation with respect to $\mathcal{I}$ and $\mathcal{R}$ for brevity.
This loss enables vector field prediction even with different timesteps for pose and image modalities.
Since our architecture conditions each modality's module on its own timestep embedding, implementing asynchronous timesteps simply involves providing separate timestep inputs to each module as $t_\mathcal{R}$ and $t_\mathcal{I}$ as in Fig.~\ref{fig:method} (a).

\setlength{\intextsep}{1pt}%
\setlength{\columnsep}{4pt}%
\begin{wrapfigure}{r}{0.4\linewidth} 
  \centering
  \includegraphics[width=\linewidth]{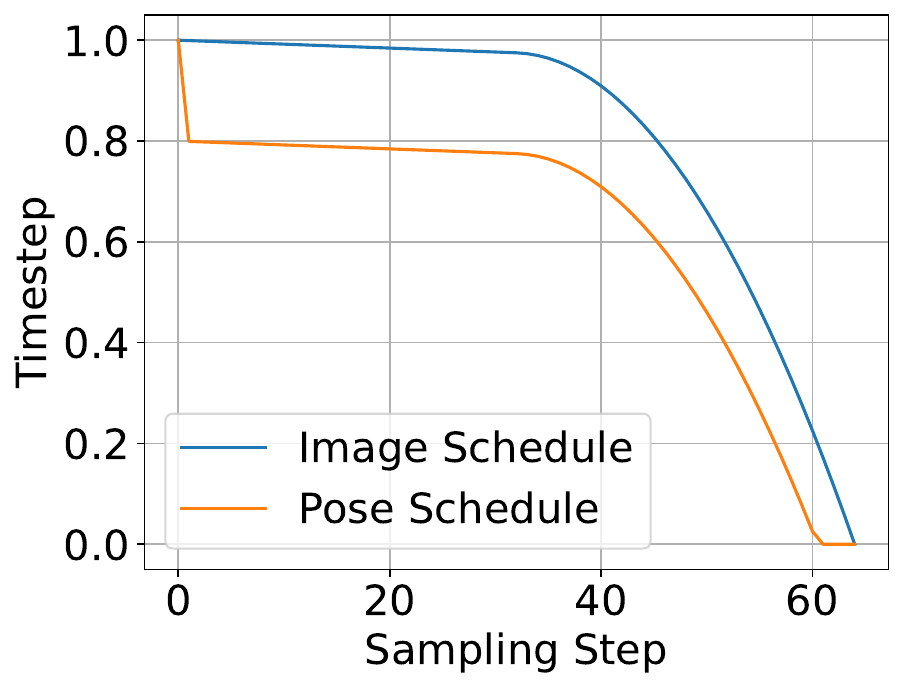}
  \vspace{-0.7cm}
  \caption{\textbf{Asynchrnous schedule ($\delta=0.2$).}}
  \label{fig:async_timestep}
\end{wrapfigure}

During sampling, we denoise the pose modality faster than images, as it is robust to fast denoising.
In implementation, both modalities start from Gaussian noise, with the pose's sampling timestep adjusted as $t_\mathcal{R} = \text{max}(t_\mathcal{I} - \delta, 0)$ as shown in Fig.~\ref{fig:async_timestep}.
However, this design introduces two potential risks: (1) excessively rapid denoising of the pose modality at early stages may introduce instability, and (2) since the pose modality reaches timestep $0$ faster, it must complete generation in fewer sampling steps.
Despite these concerns, we find that the pose modality remains robust across a wide range of $\delta$.
We hypothesize that this robustness stems from its fixed intrinsic dimensionality compared to images, which is also overparameterized as ray representations.
Thus, asynchronous sampling effectively mitigates mutual ambiguity in joint generation while avoiding these issues.

Additionally, the faster-denoised pose can more effectively guide multi-view image generation by adapting CFG~\cite{ho2022classifier} within our asynchronous framework. 
Specifically, from the viewpoint of each modality, when the opposite modality's timestep is close to $1$, its state approximates a Gaussian noise distribution, thus behaving as an unconditional modality~\cite{bao2022analytic, bao2023one}.
For instance, when $t_\mathcal{R}\approx1$, we have $\mathbb{E}[u_t^\mathcal{I} | \mathcal{I}_t, \mathcal{R}_1] \approx \mathbb{E}[u_t^\mathcal{I} | \mathcal{I}_t]$~\cite{bao2022analytic, bao2023one}.
Leveraging this property, we can adapt CFG with strength $s$ to incorporate an unconditional term for the camera rays as follows:
\vspace{\aboveeqmargin}
\vspace{-0.0cm}
\begin{equation}
    (1 + s)u_\theta(\mathcal{I}_{t_{\mathcal{I}}}, \mathcal{R}_{t_{\mathcal{R}}}, c, t_{\mathcal{I}}, t_{\mathcal{R}}) - s u_\theta(\mathcal{I}_{t_{\mathcal{I}}}, \mathcal{R}_{1}, c, t_{\mathcal{I}}, 1).
\vspace{\beloweqmargin}
\vspace{-0.6cm}
\end{equation}
\vspace{\paramargin}
\paragraph{Camera Conditioned Generation.}
Conversely, when the opposite modality’s timestep approaches $0$, the generation becomes conditioned on the opposite modality.
For instance, when $t_{\mathcal{R}}\approx0$, the conditional vector field for images becomes $\mathbb{E}[u_t^\mathcal{I}|\mathcal{I}_t,\mathcal{R}_0]$, allowing generation of camera-conditioned images. Although our main task is text-to-3DGS generation, we will demonstrate that our model naturally supports camera-conditioned multi-view image generation.

\subsection{Gaussian Splat Decoder}
\label{sec:GSdecoder}
With the dual-stream pose-video joint model presented in the previous section, the remaining task is to decode the generated latents into 3DGS representation.
To accomplish this, we employ a Gaussian Splat decoder, a 3D CNN-based architecture with attention layers, which takes the generated $\mathcal{I}$ and $\mathcal{R}$ as input, and outputs a pixel-aligned 3D Gaussian Splatting representation $\{\mu_j, \alpha_j, \Sigma_j, c_j\}_{j=1}^{H\times W\times K}$~\cite{charatan2024pixelsplat, chen2024mvsplat}.
Similar to feed-forward 3DGS methods~\cite{charatan2024pixelsplat, chen2024mvsplat, chen2024lara, szymanowicz2024flash3d, chen2024pref3r}, our Gaussian Splat decoder learns to decode 3DGS in a data-driven way, enabling fast reconstruction.

We train the Gaussian Splat decoder by rendering the decoded 3DGS to novel target views and minimizing a combination of L1 and LPIPS~\cite{zhang2018unreasonable} losses against ground-truth images, weighted as
$\mathcal{L}_{\text{target}} = \mathcal{L}_\text{L1} + 0.1\,\mathcal{L}_\text{LPIPS}.$
Furthermore, to support camera-conditioned generation, we incorporate a reconstruction objective for the source views using the same weighted combination of L1 and LPIPS losses
$\mathcal{L}_{\text{source}} = \mathcal{L}_\text{L1} + 0.1\,\mathcal{L}_\text{LPIPS}.$
Further details on Gaussian Splat Decoder are provided in Appendix~\textcolor{iccvblue}{A}.

\section{Experimental Results}

Here, we demonstrate the effectiveness of VideoRFSplat for text-to-3DGS generation. 
Our primary result is that VideoRFSplat, without SDS optimization, outperforms previous direct text-to-3DGS methods that employ SDS optimization. 

\begin{figure*}[t!]
    \centering
    \setlength\tabcolsep{2pt}
    
    \begin{tabular}
    {c
    :c@{\hspace{0.4pt}}c
    :c@{\hspace{0.4pt}}c
    :c@{\hspace{0.4pt}}c
    }
        \multirow{2}{*}{%
            \parbox{0.13\linewidth}{%
                \vspace{-8pt}%
                \centering
                \footnotesize
                \textit{A traditional wooden gate with red lanterns.}%
                    }%
        } 
        &
        \adjincludegraphics[clip,width=0.105\linewidth,trim={0 0 0 0}]{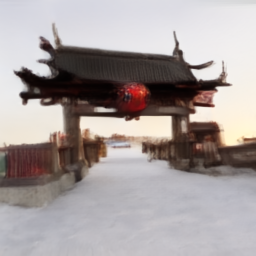} &
        \adjincludegraphics[clip,width=0.105\linewidth,trim={0 0 0 0}]{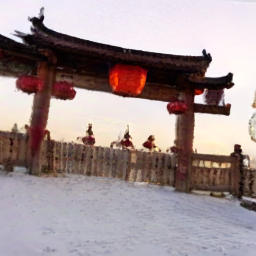} &
        
          \adjincludegraphics[clip,width=0.105\linewidth,trim={0 0 0 0}]{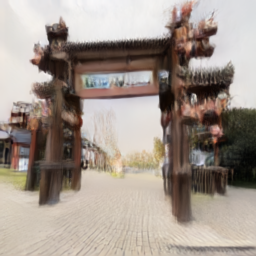} &
        \adjincludegraphics[clip,width=0.105\linewidth,trim={0 0 0 0}]{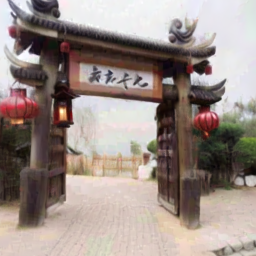} &
        
        \adjincludegraphics[clip,height=0.105\linewidth,trim={0 0 0 0}]{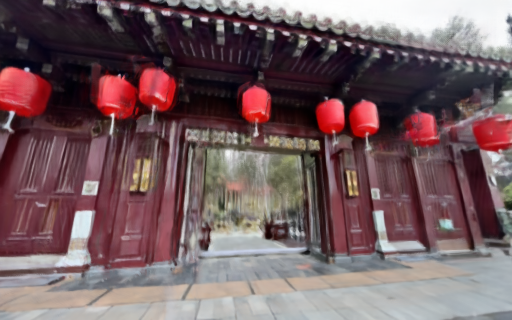} &
        \adjincludegraphics[clip,height=0.105\linewidth,trim={0 0 0 0}]{
        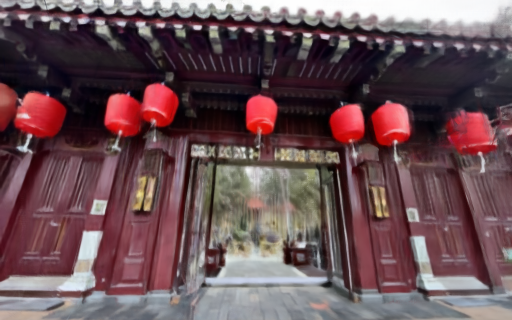
        } \\ [-2.6pt]

        & 
        \adjincludegraphics[clip,width=0.105\linewidth,trim={0 0 0 0}]{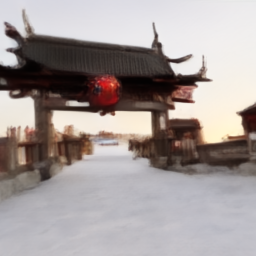} &
        \adjincludegraphics[clip,width=0.105\linewidth,trim={0 0 0 0}]{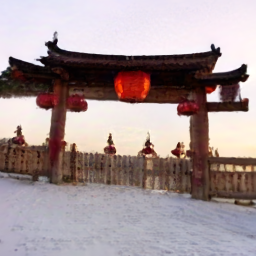} &
        
          \adjincludegraphics[clip,width=0.105\linewidth,trim={0 0 0 0}]{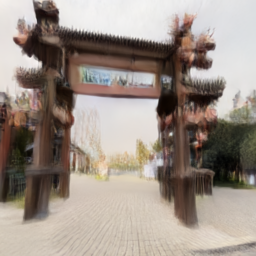} &
        \adjincludegraphics[clip,width=0.105\linewidth,trim={0 0 0 0}]{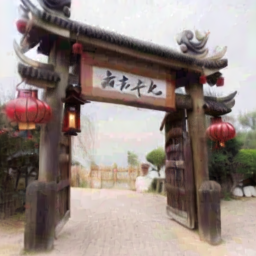} &
        
        \adjincludegraphics[clip,height=0.105\linewidth,trim={0 0 0 0}]{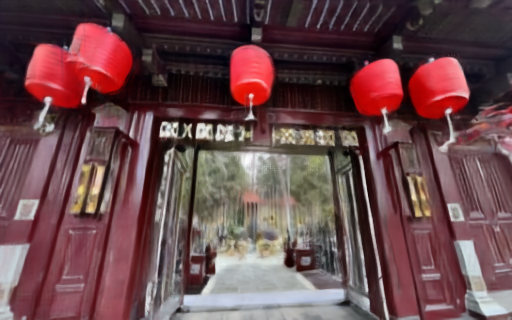} &
        \adjincludegraphics[clip,height=0.105\linewidth,trim={0 0 0 0}]{
        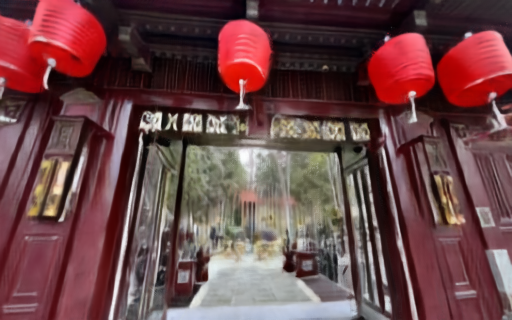} \\  [-2.6pt]
        \midrule
        \multirow{2}{*}{%
            \parbox{0.13\linewidth}{%
                \vspace{-20pt}%
                \centering
                \footnotesize
                \textit{A large stone fountain surrounded by lush greenery and a clear blue sky.}%
                    }%
        } 
        &
        \adjincludegraphics[clip,width=0.105\linewidth,trim={0 0 0 0}]{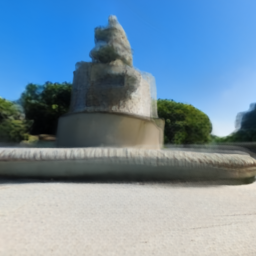} &
        \adjincludegraphics[clip,width=0.105\linewidth,trim={0 0 0 0}]{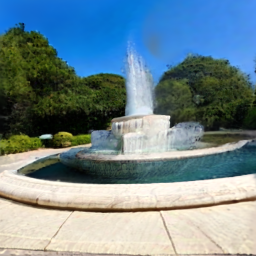} &
        
          \adjincludegraphics[clip,width=0.105\linewidth,trim={0 0 0 0}]{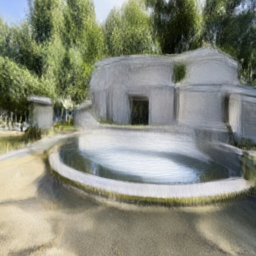} &
        \adjincludegraphics[clip,width=0.105\linewidth,trim={0 0 0 0}]{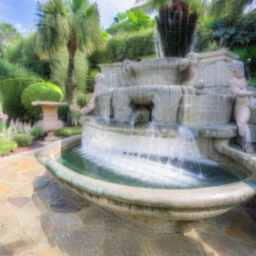} &
        
        \adjincludegraphics[clip,height=0.105\linewidth,trim={0 0 0 0}]{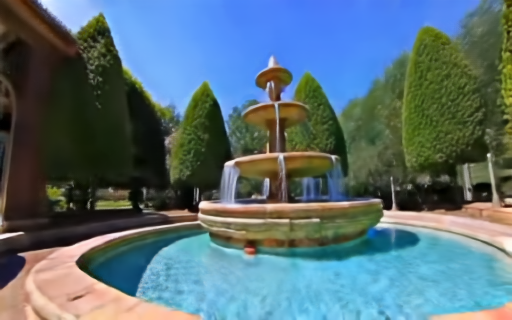} &
        \adjincludegraphics[clip,height=0.105\linewidth,trim={0 0 0 0}]{
        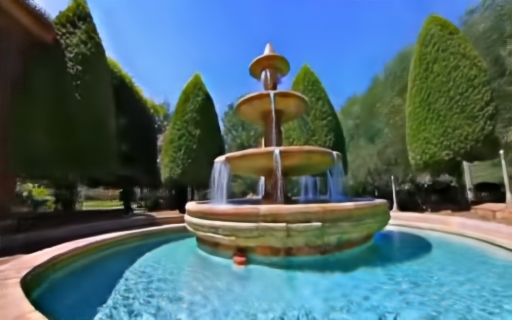
        } \\ [-2.6pt]

        & 
        \adjincludegraphics[clip,width=0.105\linewidth,trim={0 0 0 0}]{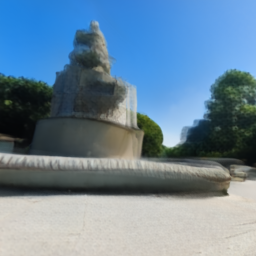} &
        \adjincludegraphics[clip,width=0.105\linewidth,trim={0 0 0 0}]{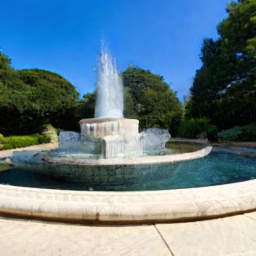} &
        
          \adjincludegraphics[clip,width=0.105\linewidth,trim={0 0 0 0}]{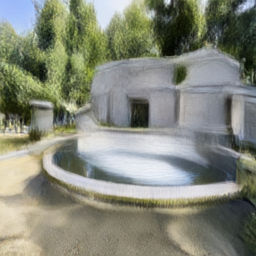} &
        \adjincludegraphics[clip,width=0.105\linewidth,trim={0 0 0 0}]{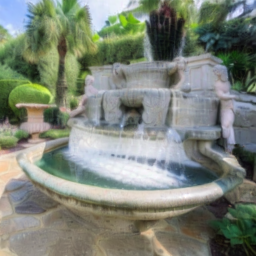} &
        
        \adjincludegraphics[clip,height=0.105\linewidth,trim={0 0 0 0}]{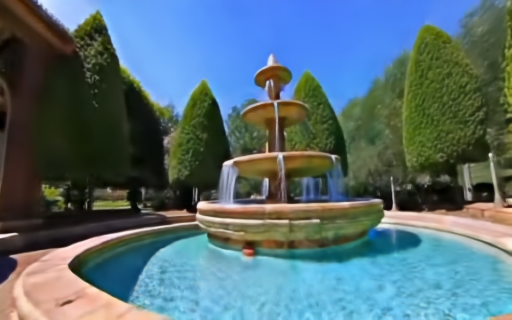} &
        \adjincludegraphics[clip,height=0.105\linewidth,trim={0 0 0 0}]{
        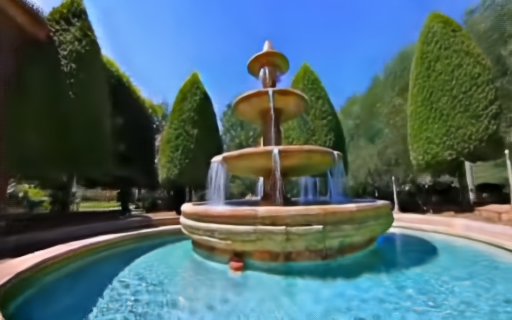} \\  [-2.6pt]
        \midrule

        \multirow{2}{*}{%
            \parbox{0.13\linewidth}{%
                \vspace{-15pt}%
                \centering
                \footnotesize
                \textit{A pair of cowboy boots in a barn.}%
                    }%
        } 
        &
        \adjincludegraphics[clip,width=0.105\linewidth,trim={0 0 0 0}]{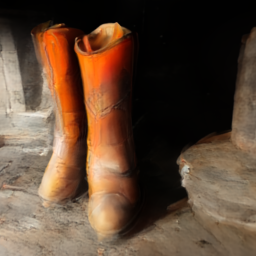} &
        \adjincludegraphics[clip,width=0.105\linewidth,trim={0 0 0 0}]{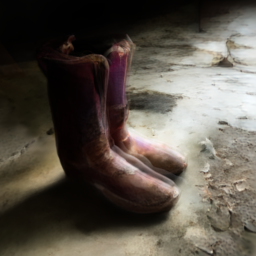} &
        
          \adjincludegraphics[clip,width=0.105\linewidth,trim={0 0 0 0}]{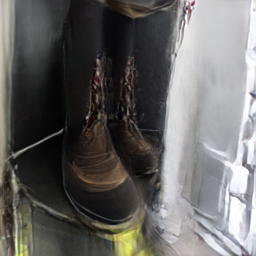} &
        \adjincludegraphics[clip,width=0.105\linewidth,trim={0 0 0 0}]{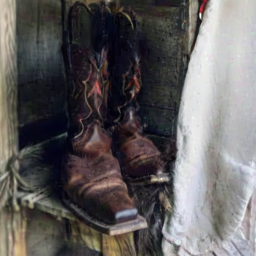} &
        
        \adjincludegraphics[clip,height=0.105\linewidth,trim={0 0 0 0}]{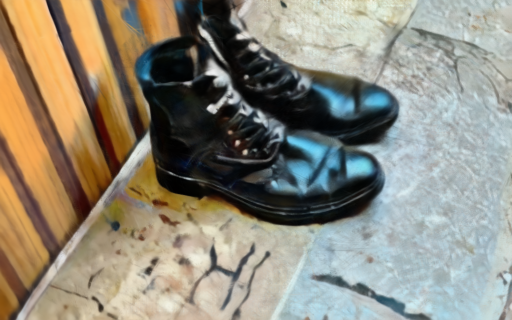} &
        \adjincludegraphics[clip,height=0.105\linewidth,trim={0 0 0 0}]{
        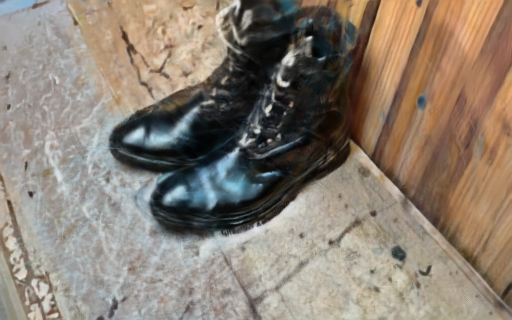
        } \\ [-2.6pt]

        & 
        \adjincludegraphics[clip,width=0.105\linewidth,trim={0 0 0 0}]{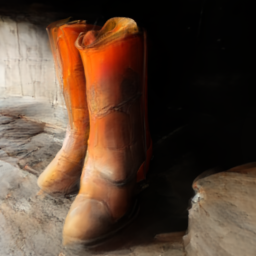} &
        \adjincludegraphics[clip,width=0.105\linewidth,trim={0 0 0 0}]{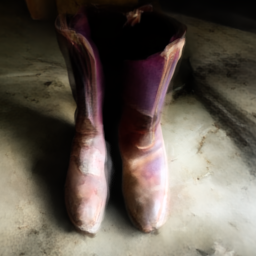} &
        
          \adjincludegraphics[clip,width=0.105\linewidth,trim={0 0 0 0}]{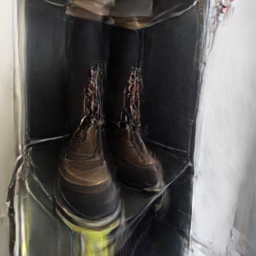} &
        \adjincludegraphics[clip,width=0.105\linewidth,trim={0 0 0 0}]{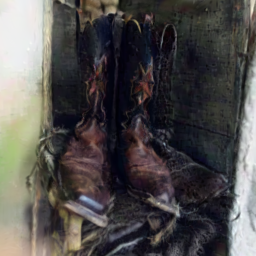} &
        
        \adjincludegraphics[clip,height=0.105\linewidth,trim={0 0 0 0}]{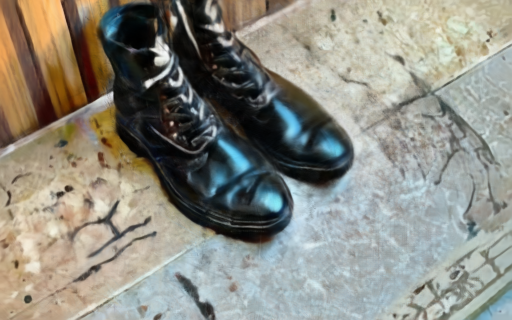} &
        \adjincludegraphics[clip,height=0.105\linewidth,trim={0 0 0 0}]{
        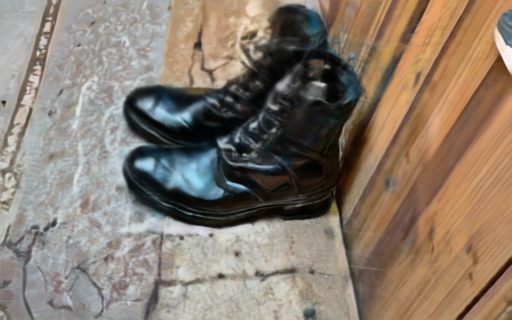} \\  [-2.6pt]
        \midrule
        \multirow{2}{*}{%
            \parbox{0.13\linewidth}{%
                \vspace{-5pt}%
                \centering
                \footnotesize
                \textit{Light blue computer mouse with green light .}%
                    }%
        } 
        &
        \adjincludegraphics[clip,width=0.105\linewidth,trim={0 0 0 0}]{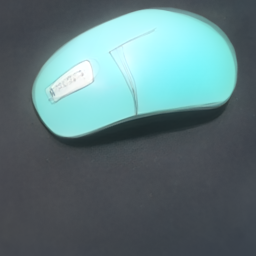} &
        \adjincludegraphics[clip,width=0.105\linewidth,trim={0 0 0 0}]{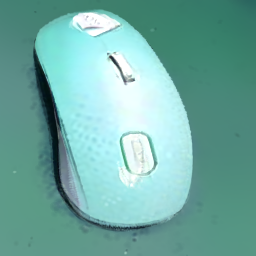} &
          \adjincludegraphics[clip,width=0.105\linewidth,trim={0 0 0 0}]{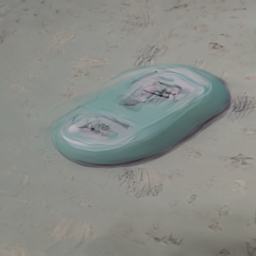} &
        \adjincludegraphics[clip,width=0.105\linewidth,trim={0 0 0 0}]{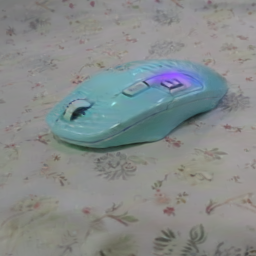} &
        \adjincludegraphics[clip,height=0.105\linewidth,trim={0 0 0 0}]{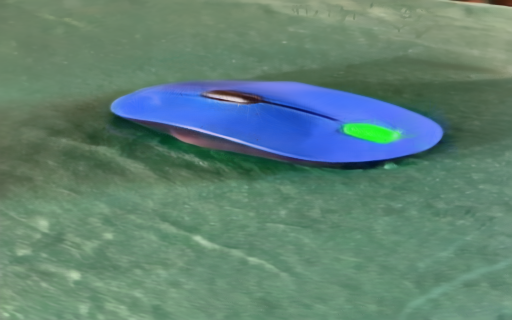} &
        \adjincludegraphics[clip,height=0.105\linewidth,trim={0 0 0 0}]{
        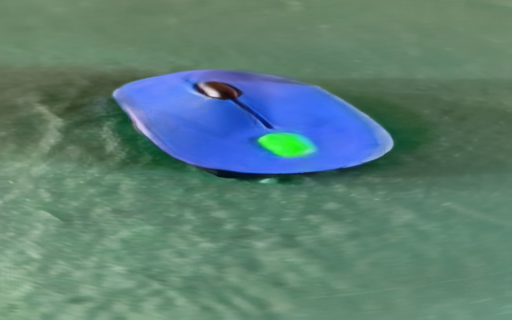
        } \\ [-2.6pt]

        & 
        \adjincludegraphics[clip,width=0.105\linewidth,trim={0 0 0 0}]{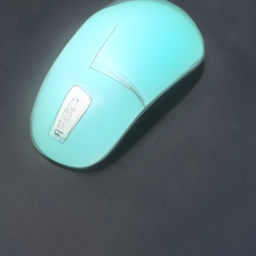} &
        \adjincludegraphics[clip,width=0.105\linewidth,trim={0 0 0 0}]{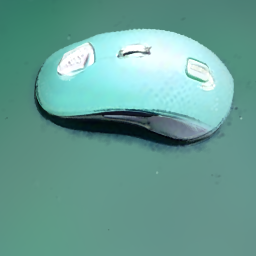} &
        
          \adjincludegraphics[clip,width=0.105\linewidth,trim={0 0 0 0}]{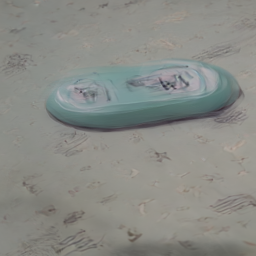} &
        \adjincludegraphics[clip,width=0.105\linewidth,trim={0 0 0 0}]{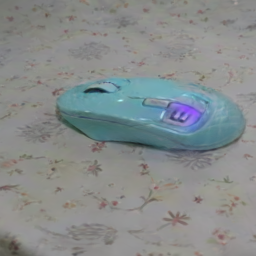} &
        \adjincludegraphics[clip,height=0.105\linewidth,trim={0 0 0 0}]{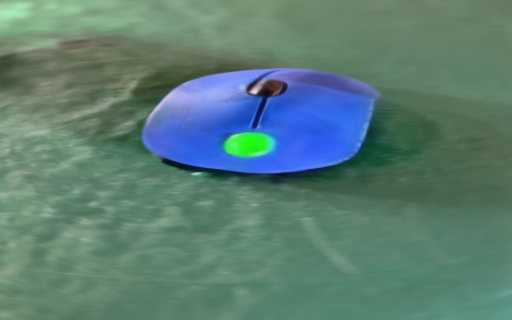} &
        \adjincludegraphics[clip,height=0.105\linewidth,trim={0 0 0 0}]{
        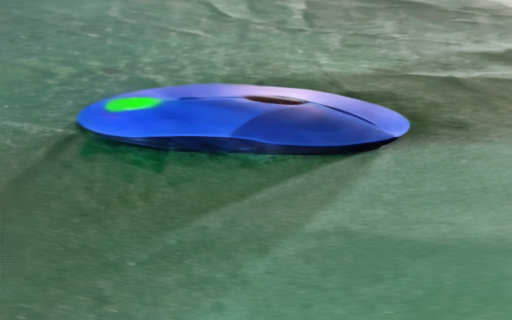} \\  [-2.6pt]

        & {\footnotesize Original} & {\footnotesize w/ SDS++} & {\footnotesize Original} & {\footnotesize w/ SDS++} & \multicolumn{2}{c}{\footnotesize Original}
        \\  [-1.0pt]
        \multicolumn{1}{c}{\small Text} & \multicolumn{2}{c}{\small Director3D~\cite{li2025director3d}} & \multicolumn{2}{c}{\small SplatFlow~\cite{go2024splatflow}} & \multicolumn{2}{c}{\small Ours}
    \end{tabular}
    \vspace{-0.2cm}
    \caption{
    \textbf{Qualitative comparison of text-to-3DGS generation on DL3DV~\cite{ling2024dl3dv} and MVImgNet~\cite{yu2023mvimgnet} validation sets as well as T3Bench~\cite{he2023t}.}
    Rendered scenes: First two rows from DL3DV, the third row from T3Bench, and the last row from MVImgNet. 
    Despite not using SDS++~\cite{li2025director3d}, VideoRFSplat generates detailed, visually consistent scenes, producing appropriate scene-specific camera poses.
    }
    \vspace{-4mm}
    \label{fig:qual_scene_generation}
\end{figure*}

\subsection{Experimental Setups}
\label{sec:Experimental_setups}

Due to space constraints, we present a brief overview of the experimental setup.
Comprehensive details of all experimental setups are available in Appendix~\textcolor{iccvblue}{B}.

\vspace{-0.1cm}
\vspace{\paramargin}
\paragraph{Datasets and Training.}
We utilize four real-world datasets for training: RealEstate10K~\cite{zhou2018stereo}, MVImgNet~\cite{yu2023mvimgnet}, DL3DV-10K~\cite{ling2024dl3dv}, and ACID~\cite{liu2021infinite}.
Since these datasets lack paired textual descriptions, we generate textual annotation using InternVL-2.5-26B~\cite{chen2024expanding}, producing multiple captions per sequence. 
We will release these annotations to support future research.  
For the video generation backbone, we use Mochi~\cite{genmo2024mochi}.
We resize all frames to $320\times512$ during training, setting $K=8$, as in SplatFlow~\cite{go2024splatflow} and Director3D~\cite{li2025director3d}.  

\vspace{-0.2cm}
\vspace{\paramargin}
\paragraph{Baselines.}
To our knowledge, SplatFlow~\cite{go2024splatflow} and Director3D~\cite{li2025director3d} are current main competitors among available direct text-to-3DGS generation methods.
As both methods use SDS++~\cite{li2025director3d} as a refinement step, we compare two variants for each method: with and without SDS++.
Additionally, for evaluations on T3Bench~\cite{he2023t}, we include previously reported baseline results~\cite{go2024splatflow, li2025director3d}.

\vspace{-0.2cm}
\vspace{\paramargin}
\paragraph{Evaluation Protocol.}
Following previous works~\cite{go2024splatflow, li2025director3d}, we evaluate our model on the MVImgNet and DL3DV validation datasets, as well as the T3Bench benchmark~\cite{he2023t}.
MVImgNet and DL3DV serve as in-domain tests for our method, Director3D, and SplatFlow, while T3Bench tests out-of-domain prompts.  
As in~\cite{go2024splatflow}, we use FID~\cite{heusel2017gans} to assess image quality and CLIP score for text-image alignment on the MVImgNet and DL3DV validation sets.
For T3Bench evaluations, we adopt established benchmark metrics, including CLIP score~\cite{hessel2021clipscore}, NIQE~\cite{mittal2012making}, and BRISQUE~\cite{mittal2012no}.


\subsection{Main Results: Text-to-3DGS Generation}
\begin{table}[t]
    \centering
    \setlength\tabcolsep{3pt}
    \resizebox{0.95\linewidth}{!}{
    \begin{tabular}{lccc}
       \toprule
        Method   & BRISQUE$\downarrow$ & NIQE$\downarrow$ & CLIPScore$\uparrow$ \\
       \midrule
       DreamFusion~\cite{poole2023dreamfusion} & 90.2 & 10.48 & -   \\
       Magic3D~\cite{lin2023magic3d} & 92.8 & 11.20 & -   \\
       LatentNeRF~\cite{metzer2023latent} & 88.6 & 9.19 & - \\
       SJC~\cite{wang2023score} & 82.0 & 10.15 & - \\
       Fantasia3D~\cite{chen2023fantasia3d} & 69.6 & 7.65 & - \\
       ProlificDreamer~\cite{wang2023prolificdreamer} & 61.5 & 7.07 & - \\
       \arrayrulecolor{gray}\midrule
       Director3D~\cite{li2025director3d} & 37.1 & 6.41 & 32.0 \\
       Director3D (w/ SDS++)~\cite{li2025director3d} & 32.3 & 4.35 & 32.9 \\
       SplatFlow~\cite{go2024splatflow} & 35.6 & 5.88 & 28.9 \\
      SplatFlow (w/ SDS++)~\cite{go2024splatflow} & 32.4 & 4.24 & 33.2  \\
       \midrule
        \rowcolor{gray!25} \textbf{VideoRFSplat} & \textbf{32.2} & \textbf{4.20} & \textbf{33.6}\\
       \arrayrulecolor{black}\bottomrule
    \end{tabular}
    }
    \vspace{-0.055cm}
    \vspace{\abovetabcapmargin}
    \caption{\textbf{Quantitative results on T3Bench~\cite{he2023t}.}
    VideoRFSplat outperforms all baselines without SDS++ refinement.}
    \vspace{\belowtabcapmargin}
    \label{tab:t3bench}
\end{table}

\begin{table}[t!]
    \centering
    \setlength\tabcolsep{2.5pt}
    \resizebox{0.95\linewidth}{!}{
    \begin{tabular}{lcccc}
       \toprule
       \multirow{2}{*}{Method} & \multicolumn{2}{c}{MVImgNet~\cite{yu2023mvimgnet}} & \multicolumn{2}{c}{DL3DV~\cite{ling2024dl3dv}} \\
       \arrayrulecolor{gray}\cmidrule(lr){2-3}\cmidrule(lr){4-5} 
                               & FID-10K$\downarrow$ & CLIPScore$\uparrow$  & FID-2.4K$\downarrow$ & CLIPScore$\uparrow$  \\
       \midrule
       Director3D~\cite{li2025director3d} & 39.55 & 30.48 & 88.44 & 30.04     \\ 
       Director3D (w/ SDS++)~\cite{li2025director3d} & 41.80 & 31.00 & 95.88 & 31.68    \\
       SplatFlow~\cite{go2024splatflow} & 34.85 & 31.43 & 79.91 & 30.06  \\ 
       SplatFlow (w/ SDS++)~\cite{go2024splatflow} &  35.46 & 32.30 & 85.31 & 31.90 \\
       \midrule
       \rowcolor{gray!25}\textbf{VideoRFSplat} & \textbf{30.33} & \textbf{33.0} &  \textbf{73.69} & \textbf{32.5} \\
       \arrayrulecolor{black}\bottomrule
    \end{tabular}
    }
    \vspace{-0.15cm}
    \vspace{\abovetabcapmargin}
    \caption{\textbf{Quantitative results on MVImgNet~\cite{yu2023mvimgnet} and DL3DV~\cite{ling2024dl3dv} validation sets.}
    VideoRFSplat achieves the higher performance across all metrics without SDS++ refinement.
    }
    \vspace{\belowtabcapmargin}
    \label{tab:3dgs_quant}
\end{table}

\label{sec:3dgs_generation}

\vspace{-0.1cm}
\paragraph{Quantitative results.}

Tables~\ref{tab:t3bench} and \ref{tab:3dgs_quant} present quantitative comparisons of VideoRFSplat against baselines on T3Bench, MVImgNet, and DL3DV.
VideoRFSplat consistently outperforms baselines across all metrics without relying on SDS++ refinement.
On T3Bench, VideoRFSplat achieves the lowest BRISQUE (32.2) and NIQE (4.20), indicating improved image quality.
On MVImgNet, VideoRFSplat achieves a significantly lower FID (30.33) and a higher CLIP score (33.0) than baselines.
Similarly, on DL3DV, it achieves the lowest FID (73.69) and highest CLIP score (32.5).
We attribute these strong results, achieved without SDS-based refinement, to three key factors:
\textbf{(1)} leveraging a powerful pre-trained video generation model,
\textbf{(2)} adopting a more effective architectural design, and
\textbf{(3)} incorporating an improved sampling strategy.
We analyze these factors in the following sections.

\vspace{-0.05cm}
\vspace{\paramargin}
\paragraph{Qualitative results.}

Consistent with our quantitative findings, Fig.~\ref{fig:qual_scene_generation} shows that VideoRFSplat qualitatively outperforms SplatFlow and Director3D without SDS++ refinement.
Specifically, Director3D often produces blurry outputs, lacking sharpness and detail for realism. 
Similarly, SplatFlow often struggles to render precise textures and maintain consistent fidelity.
In contrast, VideoRFSplat consistently produces sharper, more detailed, and realistic images without SDS-based refinement, at \textit{higher resolutions}.  
These results highlight VideoRFSplat’s ability to generate high-quality 3DGS well-aligned with text prompts, underscoring its effectiveness over SDS-dependent methods.
Additional qualitative results are available in Appendix~\textcolor{iccvblue}{C}.


\subsection{Analysis: Asynchronous Sampling}
\label{sec:async_time}

\begin{table}[t]
\centering
\setlength\tabcolsep{3pt}
\resizebox{\linewidth}{!}{
\begin{tabular}{lcccc}
\toprule
 & \multicolumn{2}{c}{\textbf{Quality Metrics}} 
 & \multicolumn{2}{c}{\textbf{CLIPScore}$\uparrow$} \\
\cmidrule(lr){2-3} \cmidrule(lr){4-5}
\textbf{Method} 
  & BRISQUE$\downarrow$ & NIQE$\downarrow$
  & Multi-view & Rendered \\
\midrule
Vanilla sampling                                      & \textbf{34.9} & \textbf{5.32} & \textbf{33.3} & \textbf{32.8} \\ 
Vanilla sampling (w/o Eq.~\ref{eq:timestep_loss})     & 35.3 & 5.64 & \textbf{33.3} & \textbf{32.8} \\
\midrule
$\delta=0.1$, w/o modified CFG                         & 34.0 & 4.43 & 33.8 & 33.2 \\ 
$\delta=0.2$, w/o modified CFG                         & 34.1 & 4.39 & 34.0 & 33.2 \\
$\delta=0.3$, w/o modified CFG                         & \textbf{33.9} & \textbf{4.36} & \textbf{34.0} & \textbf{33.3} \\
$\delta=0.4$, w/o modified CFG                         & 34.6 & 4.67 & 34.2 & 33.3 \\
$\delta=0.5$, w/o modified CFG                         & 35.2 & 4.9 & 34.3 & 33.1 \\
\midrule
$\delta=0.1$, Asynchronous Sampling                    & 32.8 & 4.30 & 33.8 & 33.4 \\
\rowcolor{gray!25}$\delta=0.2$, Asynchronous Sampling                    & \textbf{32.2} & \textbf{4.20} & \textbf{34.0} & \textbf{33.6} \\
$\delta=0.3$, Asynchronous Sampling                    & 33.0 & 4.27 & 33.9 & 33.5 \\
$\delta=0.4$, Asynchronous Sampling                    & 33.8 & 4.52 & 33.8 & 33.4 \\
$\delta=0.5$, Asynchronous Sampling                    & 34.4 & 4.76 & 33.8 & 33.3 \\
\midrule
Vanilla sampling - 200 steps                          & 33.8 & 4.56 & 33.7 & 33.1 \\
Async sampling - 200 steps ($\delta=0.2$)       & \textbf{31.8} & \textbf{4.19} & \textbf{34.2} & \textbf{33.7} \\
\bottomrule
\end{tabular}
}
\vspace{\abovetabcapmargin}
\caption{\textbf{Ablation study on asynchronous sampling.}
We also report CLIP scores on multi-view images to assess text alignment of not lifted images to 3DGS.
}
\vspace{\belowtabcapmargin}
\label{tab:ablation_on_asynchronous_sampling}
\end{table}
\begin{figure}[t!]
\setlength\tabcolsep{1pt}
\begin{tabular}{cccc}
\toprule
     \raisebox{0.01\linewidth}{\rotatebox{90}{\small Vanilla.}}
      \adjincludegraphics[clip,width=0.23\linewidth,trim={0 0 0 0}]{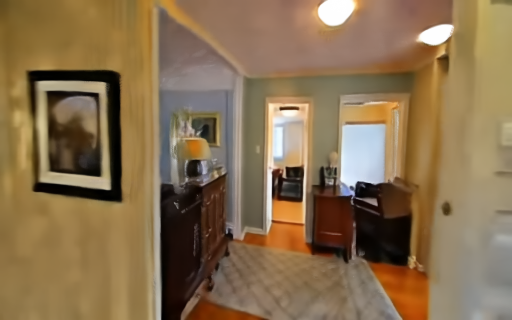} &
        \adjincludegraphics[clip,width=0.23\linewidth,trim={0 0 0 0}]{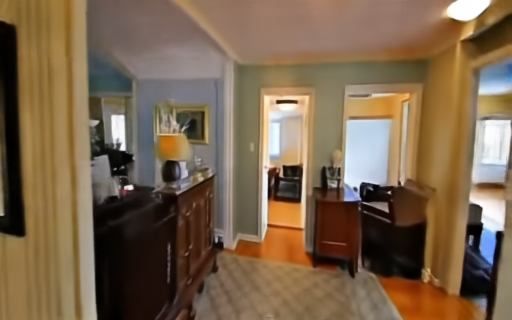} &
        \adjincludegraphics[clip,width=0.23\linewidth,trim={0 0 0 0}]{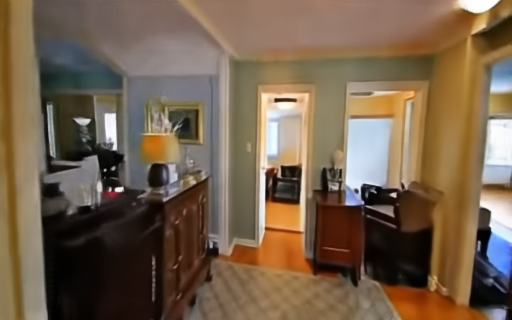} &
        \adjincludegraphics[clip,width=0.23\linewidth,trim={0 0 0 0}]{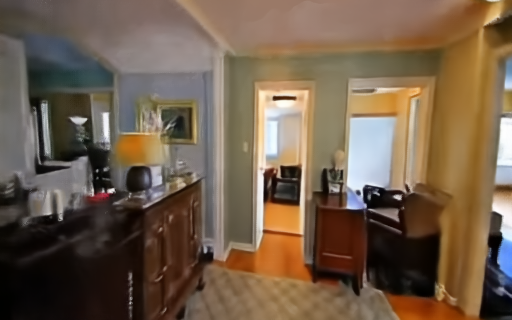}\\
     \raisebox{0.01\linewidth}{\vspace{-0.1cm}\rotatebox{90}{\small Async.}}
      \adjincludegraphics[clip,width=0.23\linewidth,trim={0 0 0 0}]{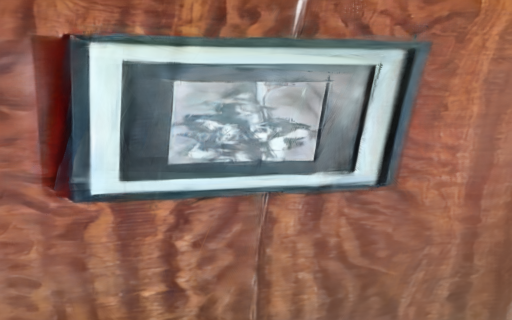} &
        \adjincludegraphics[clip,width=0.23\linewidth,trim={0 0 0 0}]{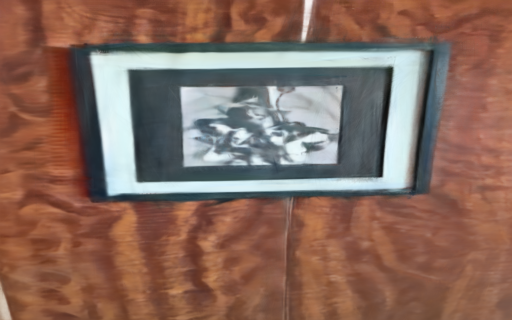} &
        \adjincludegraphics[clip,width=0.23\linewidth,trim={0 0 0 0}]{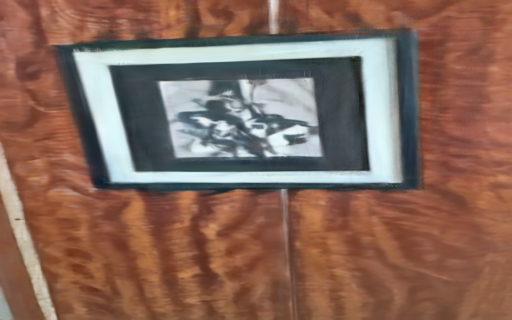} &
        \adjincludegraphics[clip,width=0.23\linewidth,trim={0 0 0 0}]{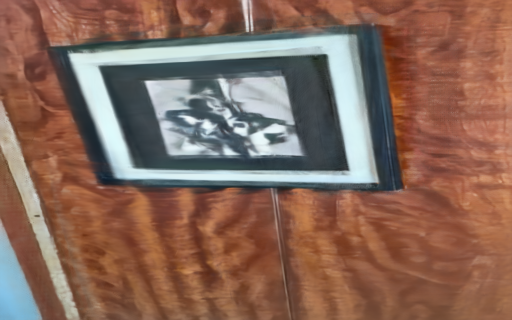}\\[-4pt]
\multicolumn{4}{p{0.98\linewidth}}{\centering\footnotesize \quad\quad \textit{"A black and white photograph framed in dark mahogany"}} \\ 
\midrule
     \raisebox{0.01\linewidth}{\rotatebox{90}{\small Vanilla.}}
      \adjincludegraphics[clip,width=0.23\linewidth,trim={0 0 0 0}]{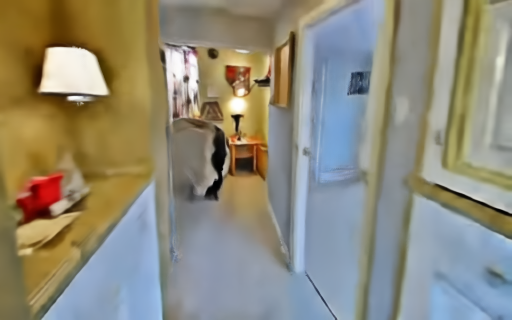} &
        \adjincludegraphics[clip,width=0.23\linewidth,trim={0 0 0 0}]{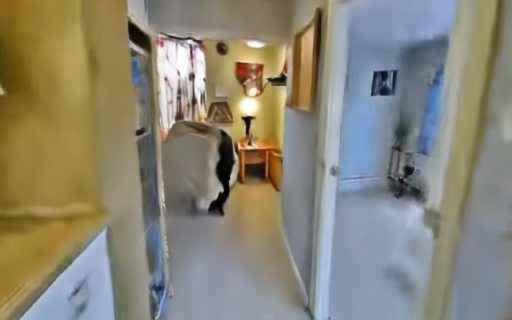} &
        \adjincludegraphics[clip,width=0.23\linewidth,trim={0 0 0 0}]{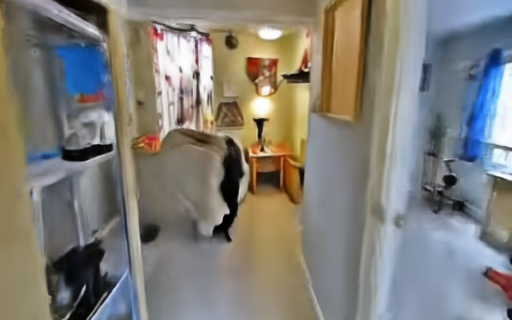} &
        \adjincludegraphics[clip,width=0.23\linewidth,trim={0 0 0 0}]{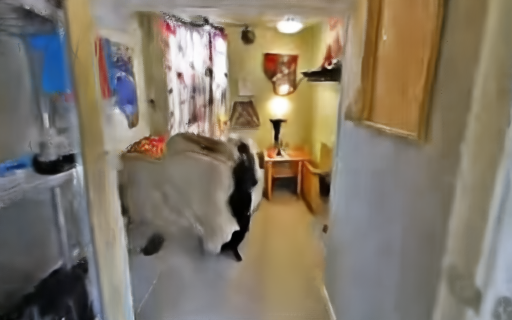}\\
     \raisebox{0.01\linewidth}{\vspace{-0.1cm}\rotatebox{90}{\small Async.}}
      \adjincludegraphics[clip,width=0.23\linewidth,trim={0 0 0 0}]{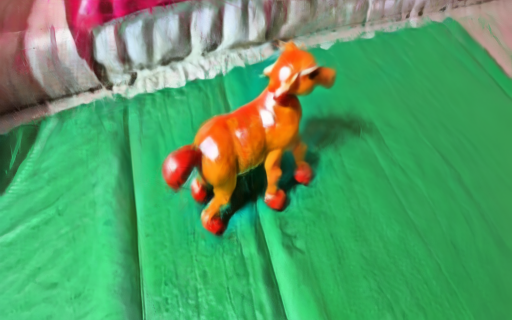} &
        \adjincludegraphics[clip,width=0.23\linewidth,trim={0 0 0 0}]{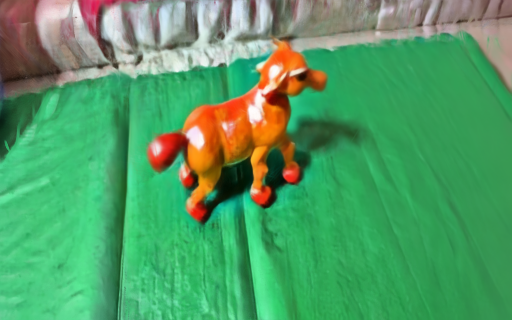} &
        \adjincludegraphics[clip,width=0.23\linewidth,trim={0 0 0 0}]{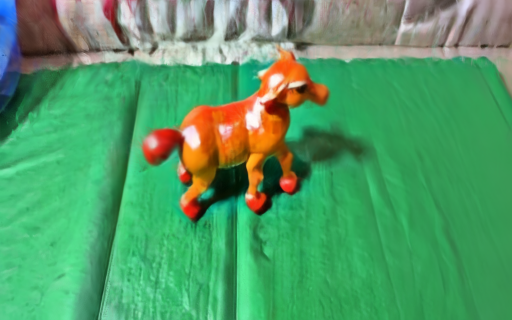} &
        \adjincludegraphics[clip,width=0.23\linewidth,trim={0 0 0 0}]{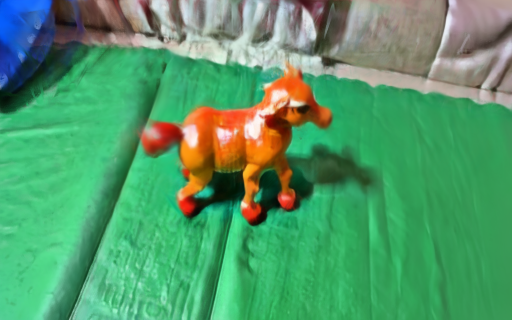}\\[-4pt]
\multicolumn{4}{p{0.98\linewidth}}{\centering\footnotesize \quad\quad \textit{"A wooden rocking horse in a child's playroom"}}\\
\bottomrule
\end{tabular}
\vspace{-0.1cm}
    \vspace{\abovefigcapmargin}
    \caption{\textbf{Effectiveness of asynchronous sampling.}
    Asynchronous sampling enhances camera poses for better key object framing, while vanilla sampling mispredicts poses, misframing key objects.
    }
    \vspace{\belowfigcapmargin}
    \vspace{-0.1cm}
    \label{fig:asynchrnous_sampling}
\end{figure}

Here, we delve into the effectiveness and characteristics of our asynchronous sampling.
In addition to metrics for T3Bench, we compute CLIP scores on generated multi-view images to assess text alignment for unlifted images to 3DGS.

\vspace{-0.05cm}
\vspace{\paramargin}
\paragraph{Effectiveness of asynchronous sampling.}
Table~\ref{tab:ablation_on_asynchronous_sampling} shows that asynchronous sampling achieves better quality metrics and CLIP scores than vanilla sampling, with peak improvements at $\delta=0.2$.
Moreover, asynchronous sampling remains robustly effective even with 200 sampling steps.  
As shown in Fig.\ref{fig:asynchrnous_sampling}, asynchronous sampling not only mitigates the divergence issues in Fig.\ref{fig:unstability} but also yields better camera trajectories that correctly frame key objects, whereas vanilla sampling mispredicts camera trajectories, resulting in misframed key objects.
We hypothesize that uncertainty in early sampling leads to unstable pose-image interactions, destabilizing camera pose generation and ultimately degrading multi-view image quality.
In contrast, asynchronous sampling mitigates this instability, resulting in more stable and improved generation.

\vspace{-0.05cm}
\vspace{\paramargin}
\paragraph{Analysis on $\delta$}
From Table~\ref{tab:ablation_on_asynchronous_sampling}, we identify three key insights:
\textbf{(1)} First, asynchronous sampling consistently performs well across all $\delta$ values tested, regardless of whether the modified CFG is applied, showing its robustness.  
Even at $\delta=0.1$, the lowest value tested, it outperforms vanilla sampling and maintains consistent gains up to $\delta=0.5$. 
\textbf{(2)} Second, without the modified CFG, multi-view CLIP scores continue to increase as $\delta$ increases, indicating that faster pose denoising reduces ambiguity and strengthens prompt alignment in multi-view images.
However, beyond $\delta=0.3$, the rendered CLIP score and quality metrics decline, showing that excessively fast pose denoising destabilizes camera poses and degrades 3DGS quality.  
\textbf{(3)} Lastly, applying modified CFG generally improves quality metrics and rendered CLIP scores compared to not using it  (except at $\delta=0.4$). 
This suggests that modified CFG better aligns with rapidly denoised camera poses, improving rendered image quality.  
Notably, unlike the case without modified CFG, increasing $\delta$ does not monotonically improve multi-view CLIP scores; instead, scores peak at $\delta=0.2$ before declining. 
Nevertheless, rendered CLIP scores remain higher than those without modified CFG.
This suggests that while the modified CFG strengthens pose-image alignment and improves 3DGS quality, excessive guidance from unstable pose at high $\delta$ may degrade prompt alignment.

\vspace{-0.05cm}
\vspace{\paramargin}
\paragraph{Effect of training with Eq.~\ref{eq:timestep_loss}}
Additionally, we evaluate performance when the training scheme does not employ timestep division as outlined in Eq.~\ref{eq:timestep_loss}. Our proposed training scheme, which divides timesteps, demonstrates slightly better results. This suggests that our approach of dividing timesteps during training is not detrimental and achieves comparable or marginally improved performance.

\subsection{Analysis: Architectural Approach}
\label{sec:architecture_approach}

\begin{figure}[t!]
\setlength\tabcolsep{1pt}
\begin{tabular}{cc:cc}
      \adjincludegraphics[clip,width=0.23\linewidth,trim={0 0 0 0}]{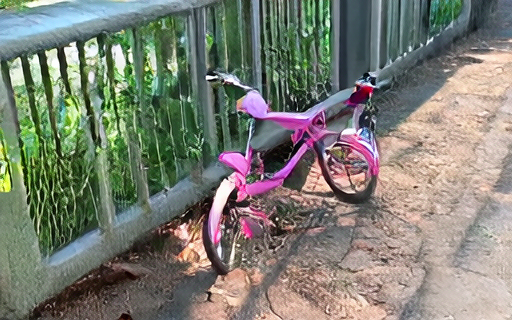} &
        \adjincludegraphics[clip,width=0.23\linewidth,trim={0 0 0 0}]{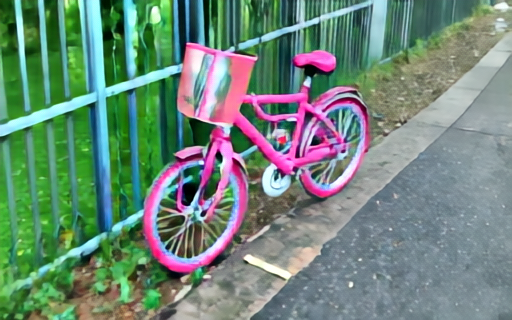} &
        \adjincludegraphics[clip,width=0.23\linewidth,trim={0 0 0 0}]{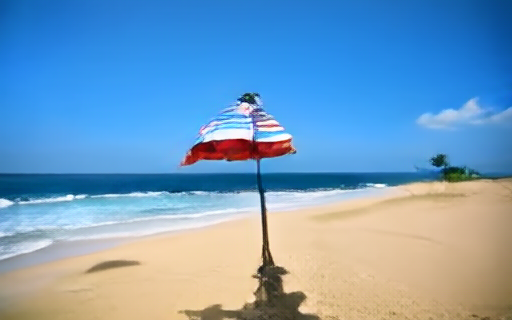} &
        \adjincludegraphics[clip,width=0.23\linewidth,trim={0 0 0 0}]{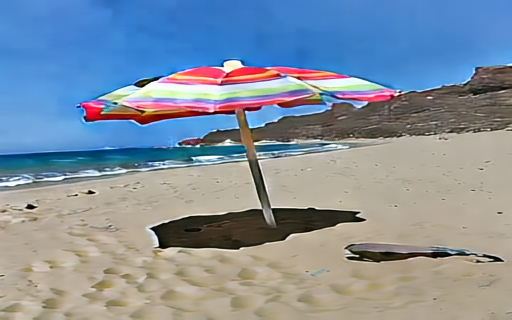}\\[-4pt]

\multicolumn{2}{p{0.48\linewidth}}{\centering\footnotesize \textit{"A pink bicycle leaning against a fence"}} &  \multicolumn{2}{p{0.48\linewidth}}{\centering\footnotesize \textit{"A striped beach umbrella standing tall on a sandy beach"}}\\ 
\end{tabular}
\vspace{-0.1cm}
    \vspace{\abovefigcapmargin}
    \caption{\textbf{Architecture Comparison.}
    For each example, \textit{Left}: channel concat architecture (SplatFlow). \textit{Right}: our architecture.
    }
    \vspace{0.1cm}
    \vspace{\belowfigcapmargin}
    \label{fig:architectural_comparison}
\end{figure}

\begin{wraptable}[4]{R}{0.6\linewidth}
    \centering
    \resizebox{\linewidth}{!}{
    \begin{tabular}{cccc}
    \toprule
         Method   & NIQE$\downarrow$ & BRISQUE$\downarrow$ & CLIPScore$\uparrow$ \\ 
         \midrule
         SplatFlow~\cite{go2024splatflow} & 7.83  & 42.09 & 29.8 \\
        \rowcolor{gray!25} Ours & \textbf{6.19} & \textbf{39.11} & \textbf{32.5} \\ 
    \bottomrule
    \end{tabular}
    }
    \vspace{-0.3cm}
    \caption{\textbf{Architecture comparison}}
    \label{tab:architecture_comparison}
\end{wraptable}

To validate the effectiveness of our architectural design, we compare it against the design scheme of SplatFlow~\cite{go2024splatflow}, which uses channel concatenation for joint pose-image generation.
We trained both models under identical conditions for 60K iterations with Mochi~\cite{genmo2024mochi} and then compared their multi-view results.  

Table~\ref{tab:architecture_comparison} shows that our architecture consistently outperforms channel concatenation of SplatFlow across all metrics.  
Our model achieves better image quality, with lower NIQE (6.19 vs. 7.83) and BRISQUE (39.11 vs. 42.09) scores, and a significantly higher CLIPScore (32.5 vs. 29.8), indicating better text alignment. 
Figure~\ref{fig:architectural_comparison} qualitatively illustrates these improvements, preserving more details of images than SplatFlow.  
This highlights that parameter-sharing like channel concatenation can degrade joint pose-image generation, underscoring the benefits of our dual-stream approach.

\subsection{Results: Camera Conditioned Generation}
\label{sec:cam_ctrl}
\begin{table}[t]
    \centering
    \setlength\tabcolsep{3pt}
    \resizebox{0.8\linewidth}{!}{
    \begin{tabular}{lcccc} 
       \toprule
       Method & FID-8K$\downarrow$ & TransErr$\downarrow$ & RotErr$\downarrow$ & CLIPScore$\uparrow$ \\
       \midrule
       MotionCtrl~\cite{wang2024motionctrl} & 69.04 & 0.109 & 0.5648 & 29.8 \\
       CameraCtrl~\cite{he2024cameractrl} & 62.97 & 0.088 & 0.5176 & 30.1 \\
       \midrule
       \rowcolor{gray!25}
       \textbf{VideoRFSplat} & \textbf{43.07} & \textbf{0.063} & \textbf{0.4223} & \textbf{31.1} \\
       \arrayrulecolor{black}\bottomrule
    \end{tabular}
    }
    \vspace{-0.2cm}
    \caption{\textbf{Results on camera conditioned generation.}
    VideoRFSplat can perform camera conditioned generation.
    }
    \vspace{-0.4cm}
    \label{tab:camctrl}
\end{table}
With the camera timestep set to 0, VideoRFSplat generates images following given camera trajectories. 
We evaluate this both quantitatively and qualitatively. For evaluation, we use 1000 sequences from RealEstate10K~\cite{zhou2018stereo} with extracted camera trajectories and captions to generate images.  
We assess pose alignment via normalized translation and rotation errors from ParticleSFM~\cite{zhao2022particlesfm}-estimated poses against ground truth.  
We compare VideoRFSplat against MotionCtrl~\cite{wang2024motionctrl} and CameraCtrl~\cite{he2024cameractrl}, reporting quantitative results in Table~\ref{tab:camctrl}.
VideoRFSplat outperforms other methods in FID-8K (43.07), translation error (0.063), rotation error (0.4223), and CLIPScore (31.1).  
These results confirm that VideoRFSplat generates images following camera trajectories.  
Qualitative results are available in the Appendix~\textcolor{iccvblue}{C}.



\section{Conclusion}


In this paper, we introduced \textit{VideoRFSplat}, a direct text-to-3DGS model for real-world scenes.
Built upon a video generative model, our approach utilizes a dual-stream architecture to minimize the interference between pose and image modalities.
Moreover, we propose asynchronous sampling, which accelerates pose denoising while enhancing CFG, leading to a more stable joint sampling and improved generation quality.
Through extensive quantitative and qualitative evaluations, VideoRFSplat achieves state-of-the-art performance in direct text-to-3DGS generation, surpassing existing baselines even without SDS refinements. 
We believe that adapting 2D generative models for joint multi-view and pose generation is crucial for real-world scene generation, and VideoRFSplat offers valuable insights into both sampling strategies and architectural design to advance this direction.

{
    \small
    \bibliographystyle{ieeenat_fullname}
    \bibliography{main}
}

\clearpage
\appendix
\section{Details of VideoRFSplat}

Here, we present additional details on VideoRFSplat, expanding on the discussion in Section~\textcolor{iccvblue}{4} to address space constraints.
We first delve into the architecture and camera optimization of the dual-stream pose-video joint model in Section~\ref{appsec:dual-stream-pose-video}, emphasizing key modifications from the Mochi video generation model.
Following that, Section~\ref{app-sec:details_gsdecoder} details the architecture of the Gaussian Splat decoder.
Lastly, in Section~\ref{app-sec:cam-conditioned_generation}, we outline the previously omitted aspects of CFG for camera-conditioned generation.

\subsection{Dual-Stream Pose-Video Joint Model}
\label{appsec:dual-stream-pose-video}

\paragraph{Architecture.}
We adopt Mochi~\cite{genmo2024mochi} as the backbone for our video generation model without any architectural modifications. Mochi utilizes an Asymmetric Diffusion Transformer architecture with full 3D attention, enabling high-fidelity video synthesis. The model consists of approximately 10 billion parameters, and it employs the T5-XXL model~\cite{raffel2020exploring} as its text encoder to extract text-conditional embeddings. For further architectural details, we refer the reader to~\cite{genmo2024mochi}.

For the pose generation module, we employ the same Asymmetric Diffusion Transformer architecture as Mochi but with a more efficient configuration. Specifically, we set the hidden size to 256, the patch size to 2, the number of attention heads to 4, and the input size of the ray embedding to $10 \times 16$. While the video generation model consists of 48 transformer blocks, the pose model is designed with 16 blocks for computational efficiency. Both models share the same text encoder. The pose model is trained from scratch.

The communication block is placed at every three blocks in the video model and at every single block in the pose model, allowing periodic information exchange between the two streams. Communication is omitted in the final layer to maintain independent final representations. To prevent significant changes in the initial output of the video model, weights and biases of the linear layer in the communication block are initialized to zero.

\vspace{-0.4cm}
\paragraph{Ray to camera parameters.}
To recover camera parameters from generated rays, we follow a slightly optimized version of the RayDiffusion approach~\cite{zhang2024cameras}, which is utilized in SplatFlow~\cite{go2024splatflow}.
In summary, the camera center is estimated by minimizing the mismatch between rays.  
Next, the projection matrix is computed using a least-squares approach and decomposed into the intrinsic matrix and rotation.  
Finally, optimization with the Adam optimizer~\cite{kingma2014adam} is applied to refine the intrinsic and rotation matrices, enforcing shared intrinsic parameters across all views. Then, we slightly refine camera poses to be consistent with source views.

\subsection{Details of Gaussian Splat Decoder}
\label{app-sec:details_gsdecoder}
The Gaussian Splat Decoder follows a 3D-CNN architecture, adopted from the decoder of Mochi~\cite{genmo2024mochi}.  
To enhance global context modeling, we add two attention layers to each residual block in the lowest layer and do not use causal 3D convolutions~\cite{yu2023language}.  
Additionally, we incorporate Plücker ray embeddings as inputs to the decoder.  
Specifically, the embeddings follow the format of LGM~\cite{tang2024lgm}, forming a 9-channel representation.  
These embeddings are transformed to match the resolution of the intermediate representations in each decoder block and are injected via additional 3D convolution layers.  
The Gaussian Splat Decoder outputs depth, opacity, RGB, rotation, and scale, forming an 11-channel output.  
To accommodate this, we introduce an additional $1 \times 1$ convolution layer in the output layer.
For 3DGS rendering, we utilize Gsplat~\cite{ye2024gsplatopensourcelibrarygaussian} library, which offers the efficient implementation.

\subsection{Details of Camera Conditioned Generation}
\label{app-sec:cam-conditioned_generation}
\paragraph{Classifier-Free Guidance details.}
To implement camera-conditioned generation, we follow the formulation presented in~\cite{bao2023one}.  
Since our generation task involves multi-view generation conditioned on both a text prompt and a camera trajectory, we decompose these two conditions within the Classifier-Free Guidance (CFG) framework like HarmonyView~\cite{woo2024harmonyview}:  
\vspace{-0.2cm}
\begin{equation}
    \begin{aligned}
        \Big[&(1 + s_c) u_\theta(\mathcal{I}_{t_{\mathcal{I}}}, \mathcal{R}_{t_{\mathcal{R}}}, c, t_{\mathcal{I}}, t_{\mathcal{R}})\\ 
        &- s_c u_\theta(\mathcal{I}_{t_{\mathcal{I}}}, \mathcal{R}_{t_{\mathcal{R}}}, c_{\text{null}}, t_{\mathcal{I}}, t_{\mathcal{R}}) \\
        &+ (1 + s_\mathcal{R}) u_\theta(\mathcal{I}_{t_{\mathcal{I}}}, \mathcal{R}_{0.05}, c, t_{\mathcal{I}}, 0.05) \\ 
        &- s_\mathcal{R} u_\theta(\mathcal{I}_{t_{\mathcal{I}}}, \mathcal{R}_{1}, c, t_{\mathcal{I}}, 1)\Big] / 2,
    \end{aligned}
\vspace{-0.3cm}
\end{equation}
where $s_\mathcal{R}$ and $s_c$ represent the guidance strengths for the camera pose and text conditions, respectively, and $c_{\text{null}}$ denotes the null text condition used in CFG. 
Also, following DiffusionForcing~\cite{chen2025diffusion}, we slightly noise the camera pose condition to the better conditional mechanism as $t_\mathcal{R}=0.05$.

\section{Further Details of Experimental Setups}

\paragraph{Training dataset.}
We use multiple datasets for training, each containing multi-view images with camera annotations.  
The MVImgNet~\cite{yu2023mvimgnet} dataset originally consisted of 219,188 scenes with camera parameters.  
After filtering out erroneous scenes, we retained approximately 200K scenes.  
For validation, 1.25K scenes were allocated for each specific task.  
The DL3DV~\cite{ling2024dl3dv} dataset, which initially contained 10K scenes, was similarly processed, with 300 sequences designated for the validation set.  
Additionally, we incorporate the RealEstate10K~\cite{zhou2018stereo} and ACID~\cite{liu2021infinite} datasets in our training pipeline. 
However, during downloading and preprocessing, approximately 20\% of the total data was lost due to filtering and quality control steps.  
These datasets collectively provide a diverse set of multi-view scenarios:  
\begin{itemize}
    \item \textbf{MVImgNet} consists of object-centric video, capturing various objects within controlled environments.  
    \item \textbf{DL3DV} primarily contains outdoor scenes, featuring complex natural landscapes and diverse conditions.  
    \item \textbf{ACID} focuses on aerial scenes, providing a wide range of viewpoints from aerial.  
    \item \textbf{RealEstate10K} comprises indoor scenes, primarily focused around residential rooms and houses.  
\end{itemize}

To generate text captions, we extract multiple captions per sequence for the DL3DV, RealEstate10K, and ACID datasets, enabling fine-grained descriptions for each sequence.  
Each sequence is divided into groups of 32 images, and captions are generated using the InternVL2.5-26B model~\cite{chen2024expanding}.  
In contrast, for the MVImgNet dataset, we generate a single caption per scene to provide a concise summary of the overall content.  
Since MVImgNet is an object-centric dataset, the main object remains consistent across all images within a sequence, making a single caption sufficient for describing the entire scene.  
To ensure that the captions accurately capture the main object while remaining concise, we randomly select one of the following three prompts:  
\begin{itemize}
    \item \textit{"Briefly describe the main object in the image, including its color and key features, in a single concise sentence."}  
    \item \textit{"Describe the main object and its surroundings in 15 words or fewer, using keywords or a short phrase."}  
    \item \textit{"Summarize the main object's color, texture, and shape in no more than 15 words, using a concise phrase."}  
\end{itemize}

\vspace{-0.4cm}
\paragraph{Training Details.}  
We train the joint pose-video model with a batch size of 16 for 120K iterations.  
We use an initial learning rate of $5 \times 10^{-5}$ with a cosine decay schedule and a warm-up period of 1000 steps.  
Training is implemented using Fully Sharded Data Parallel (FSDP) for memory efficiency, and we optimize the model using the Adam optimizer.  
For timestep sampling, we uniformly sample both pose and multi-view timesteps.  
To further improve training efficiency, text embeddings are precomputed and stored prior to training. For training the Gaussian Splat decoder, we use a batch size of 8 for 400K iterations with a learning rate of $5 \times 10^{-5}$.  
During the first 300K iterations, we render 13 target views for training, and in the final 100K iterations, we increase the number of target views to 19.

\vspace{-0.4cm}
\paragraph{Inference and Evaluation.}  
For inference, we follow the default timestep schedule used in Mochi.  
Unless otherwise specified, we use 64 sampling steps. Additionally, as shown in the ablation study, we set $\delta=0.2$ as the default.
Metrics computation and evaluation splits are configured following the setup of SplatFlow~\cite{go2024splatflow}.  
Additionally, during sampling, one possible approach for modified Classifier-Free Guidance (CFG) is to continuously sample random poses from $\mathcal{N}(0, I)$.  
However, we found that pre-sampling the poses and keeping them fixed during generation leads to improved stability.

\section{Additional Results}

\subsection{Image-First Asynchronous Sampling}  
\begin{figure}[t!]
\setlength\tabcolsep{1pt}

\begin{tabular}{cccc}
\toprule
      \adjincludegraphics[clip,width=0.23\linewidth,trim={0 0 0 0}]{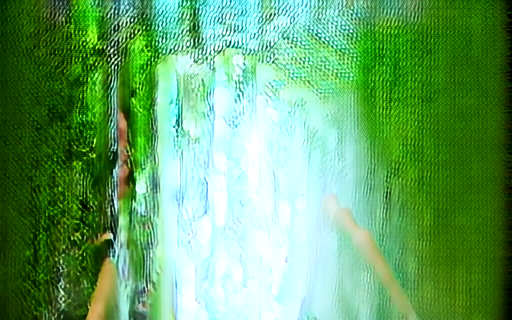} &
        \adjincludegraphics[clip,width=0.23\linewidth,trim={0 0 0 0}]{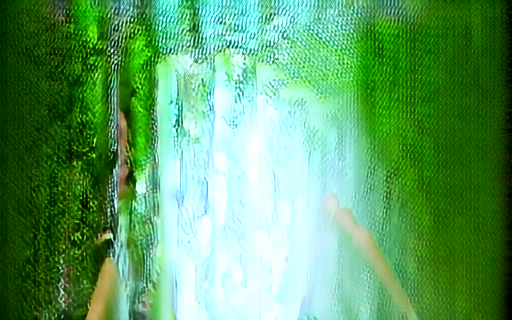} &
        \adjincludegraphics[clip,width=0.23\linewidth,trim={0 0 0 0}]{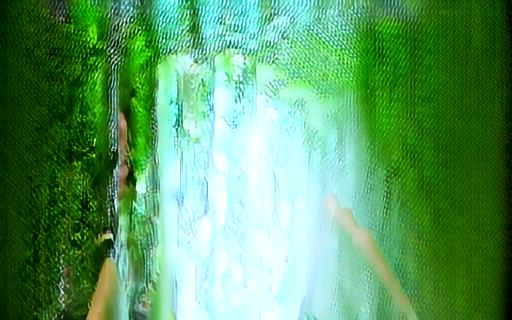} &
        \adjincludegraphics[clip,width=0.23\linewidth,trim={0 0 0 0}]{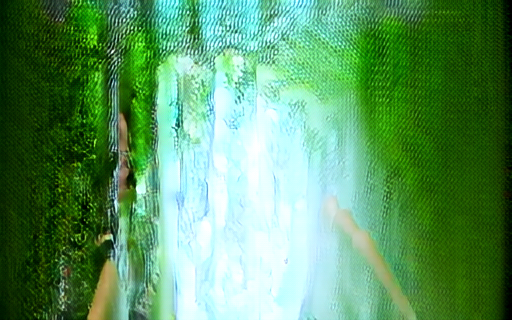}\\[-4pt]
\multicolumn{4}{p{\linewidth}}{\centering\footnotesize \textit{"A rainbow over a waterfall"}}\\ \midrule
      \adjincludegraphics[clip,width=0.23\linewidth,trim={0 0 0 0}]{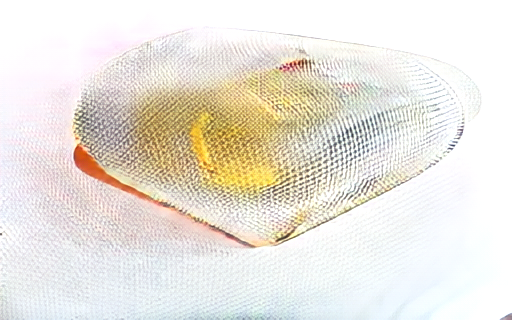} &
        \adjincludegraphics[clip,width=0.23\linewidth,trim={0 0 0 0}]{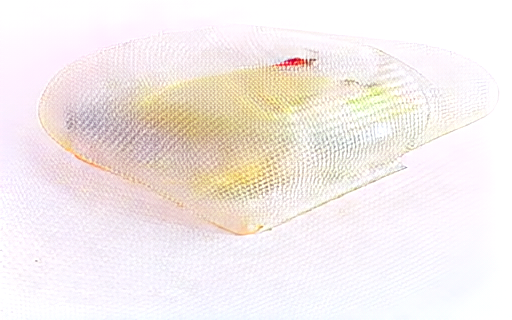} &
        \adjincludegraphics[clip,width=0.23\linewidth,trim={0 0 0 0}]{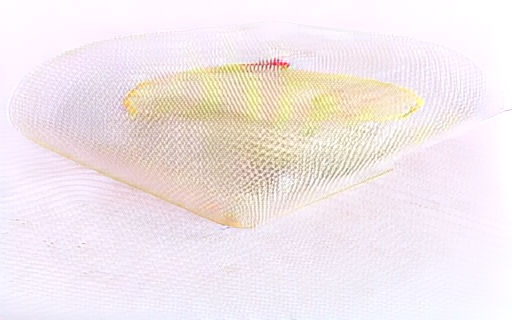} &
        \adjincludegraphics[clip,width=0.23\linewidth,trim={0 0 0 0}]{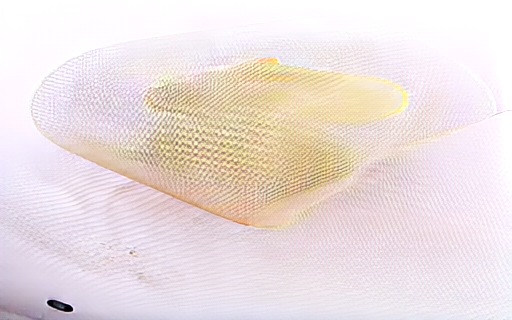}\\
\multicolumn{4}{p{\linewidth}}{\centering\footnotesize \textit{"A pair of lovebirds in a golden cage"}}\\ 
\bottomrule
\end{tabular}
    \vspace{\abovefigcapmargin}
    \caption{\textbf{Generated multi-view images from image-first asynchronous sampling.}  
    Accelerating the denoising of multi-view images, instead of the pose modality, leads to severe degeneration.}
    \vspace{\belowfigcapmargin}
    \label{fig:rgb_first}
\end{figure}

We primarily accelerate the denoising of the pose modality to reduce mutual ambiguity during the sampling process.  
However, the effectiveness of asynchronous sampling for faster multi-view images remains an open question.  
To explore this, we conduct an experiment where images' denoising is performed at an accelerated rate with $\delta = 0.2$.  
We illustrate the generated multi-view images from such image-first asynchronous sampling in Fig.~\ref{fig:rgb_first}.  
As shown in the figure, accelerating the denoising of multi-view images, rather than the pose modality, results in a significant failure in generation.  
This observation suggests that, unlike the pose modality, the image modality is not as robust to accelerated denoising, leading to severe artifacts.

\subsection{Additional Qualitative Comparison}
\begin{figure*}
    \centering
    \setlength\tabcolsep{2pt}
    
    \begin{tabular}
    {c
    :c@{\hspace{0.4pt}}c
    :c@{\hspace{0.4pt}}c
    :c@{\hspace{0.4pt}}c
    }
        \multirow{2}{*}{%
            \parbox{0.1\linewidth}{%
                \vspace{-8pt}%
                \centering
                \footnotesize
                \textit{A hot air balloon in a clear sky.}%
                    }%
        } 
        &
        \adjincludegraphics[clip,width=0.11\linewidth,trim={0 0 0 0}]{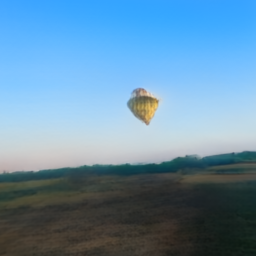} &
        \adjincludegraphics[clip,width=0.11\linewidth,trim={0 0 0 0}]{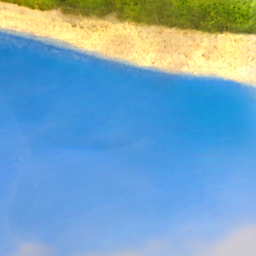} &
        
          \adjincludegraphics[clip,width=0.11\linewidth,trim={0 0 0 0}]{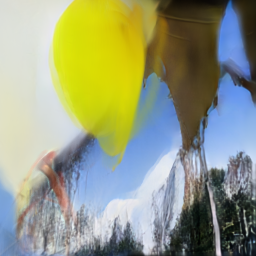} &
        \adjincludegraphics[clip,width=0.11\linewidth,trim={0 0 0 0}]{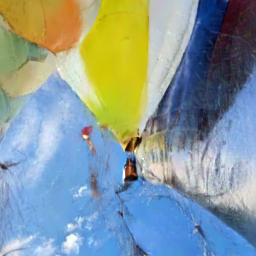} &
        
        \adjincludegraphics[clip,height=0.11\linewidth,trim={0 0 0 0}]{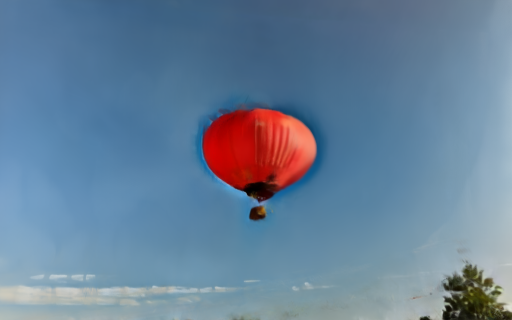} &
        \adjincludegraphics[clip,height=0.11\linewidth,trim={0 0 0 0}]{
        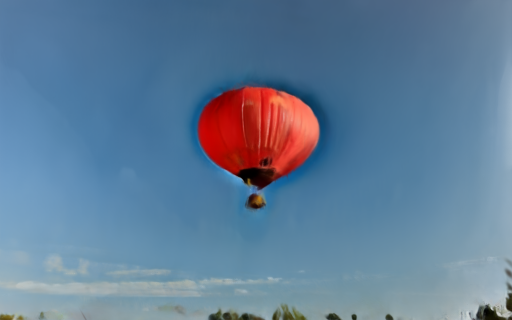
        } \\ [-2.6pt]

        & 
        \adjincludegraphics[clip,width=0.11\linewidth,trim={0 0 0 0}]{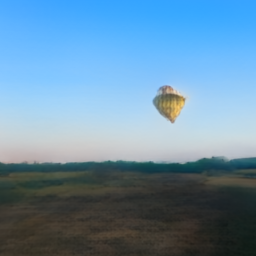} &
        \adjincludegraphics[clip,width=0.11\linewidth,trim={0 0 0 0}]{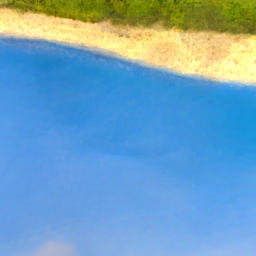} &
        
          \adjincludegraphics[clip,width=0.11\linewidth,trim={0 0 0 0}]{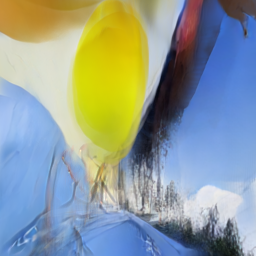} &
        \adjincludegraphics[clip,width=0.11\linewidth,trim={0 0 0 0}]{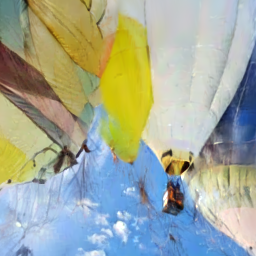} &
        
        \adjincludegraphics[clip,height=0.11\linewidth,trim={0 0 0 0}]{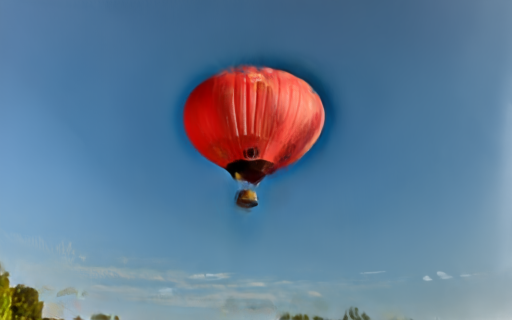} &
        \adjincludegraphics[clip,height=0.11\linewidth,trim={0 0 0 0}]{
        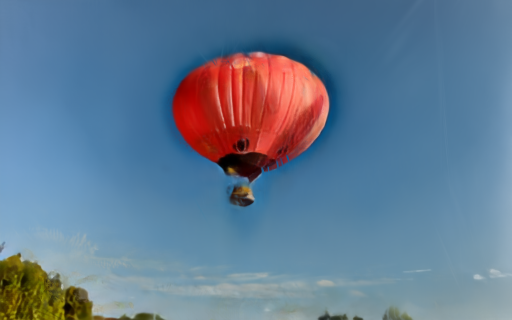} \\  [-2.6pt]
        \midrule
        \multirow{2}{*}{%
            \parbox{0.1\linewidth}{%
                \vspace{-20pt}%
                \centering
                \footnotesize
                \textit{A pair of neon running shoes waiting by the treadmill.}%
                    }%
        } 
        &
        \adjincludegraphics[clip,width=0.11\linewidth,trim={0 0 0 0}]{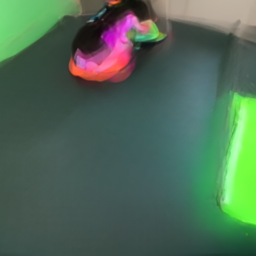} &
        \adjincludegraphics[clip,width=0.11\linewidth,trim={0 0 0 0}]{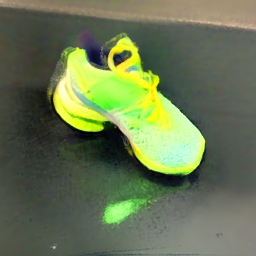} &
        
          \adjincludegraphics[clip,width=0.11\linewidth,trim={0 0 0 0}]{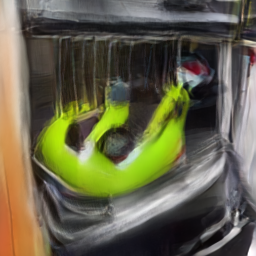} &
        \adjincludegraphics[clip,width=0.11\linewidth,trim={0 0 0 0}]{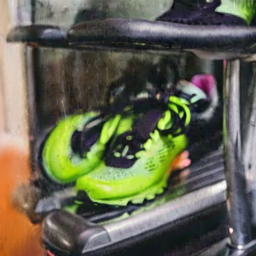} &
        
        \adjincludegraphics[clip,height=0.11\linewidth,trim={0 0 0 0}]{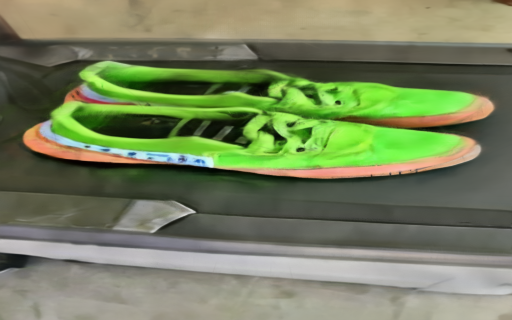} &
        \adjincludegraphics[clip,height=0.11\linewidth,trim={0 0 0 0}]{
        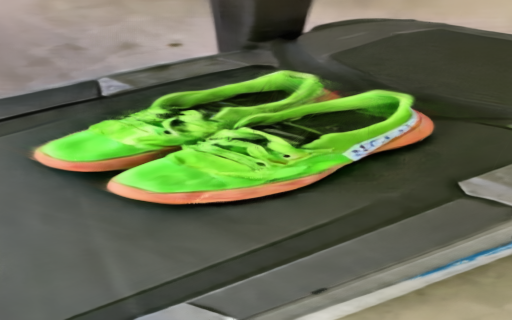
        } \\ [-2.6pt]

        & 
        \adjincludegraphics[clip,width=0.11\linewidth,trim={0 0 0 0}]{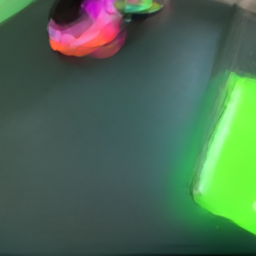} &
        \adjincludegraphics[clip,width=0.11\linewidth,trim={0 0 0 0}]{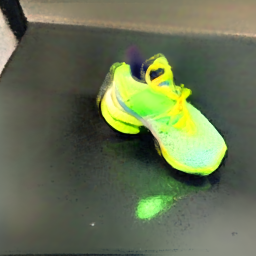} &
        
          \adjincludegraphics[clip,width=0.11\linewidth,trim={0 0 0 0}]{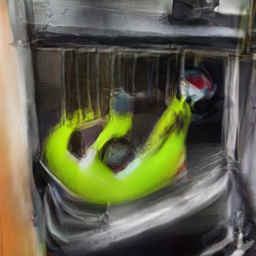} &
        \adjincludegraphics[clip,width=0.11\linewidth,trim={0 0 0 0}]{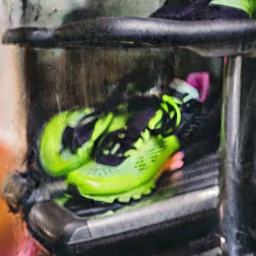} &
        
        \adjincludegraphics[clip,height=0.11\linewidth,trim={0 0 0 0}]{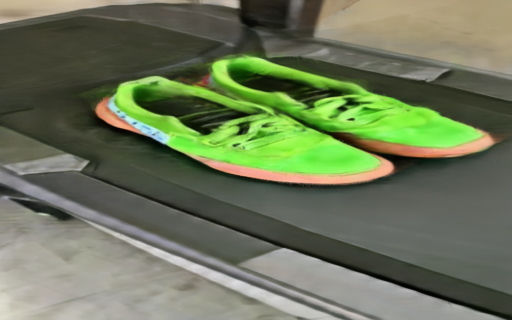} &
        \adjincludegraphics[clip,height=0.11\linewidth,trim={0 0 0 0}]{
        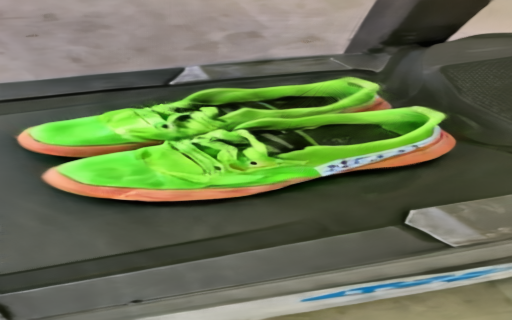} \\  [-2.6pt]
        \midrule

        \multirow{2}{*}{%
            \parbox{0.1\linewidth}{%
                \vspace{-15pt}%
                \centering
                \footnotesize
                \textit{A vintage pocket watch with a golden chain.}%
                    }%
        } 
        &
        \adjincludegraphics[clip,width=0.11\linewidth,trim={0 0 0 0}]{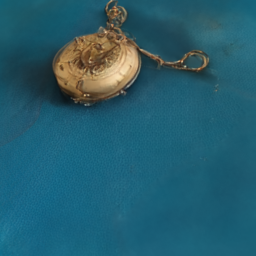} &
        \adjincludegraphics[clip,width=0.11\linewidth,trim={0 0 0 0}]{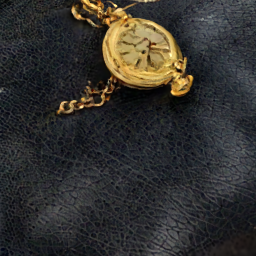} &
        
          \adjincludegraphics[clip,width=0.11\linewidth,trim={0 0 0 0}]{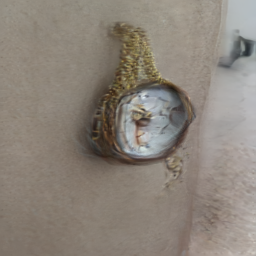} &
        \adjincludegraphics[clip,width=0.11\linewidth,trim={0 0 0 0}]{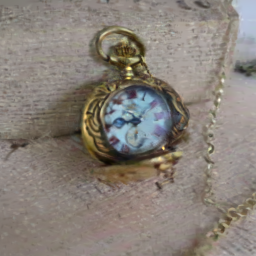} &
        
        \adjincludegraphics[clip,height=0.11\linewidth,trim={0 0 0 0}]{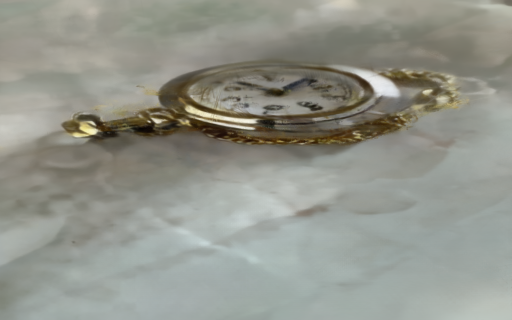} &
        \adjincludegraphics[clip,height=0.11\linewidth,trim={0 0 0 0}]{
        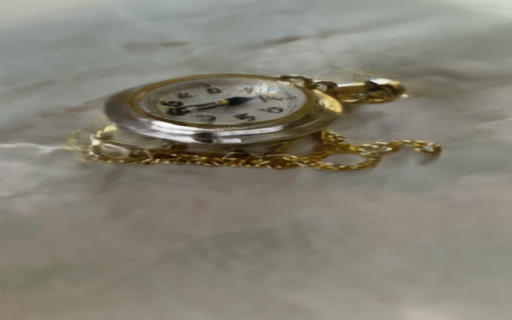
        } \\ [-2.6pt]

        & 
        \adjincludegraphics[clip,width=0.11\linewidth,trim={0 0 0 0}]{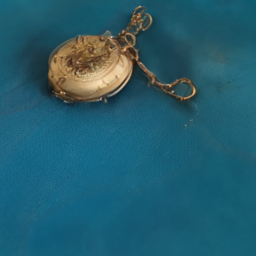} &
        \adjincludegraphics[clip,width=0.11\linewidth,trim={0 0 0 0}]{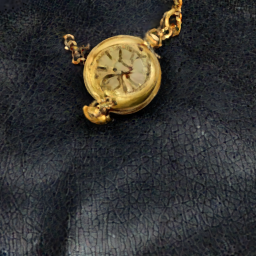} &
        
          \adjincludegraphics[clip,width=0.11\linewidth,trim={0 0 0 0}]{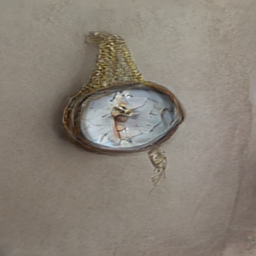} &
        \adjincludegraphics[clip,width=0.11\linewidth,trim={0 0 0 0}]{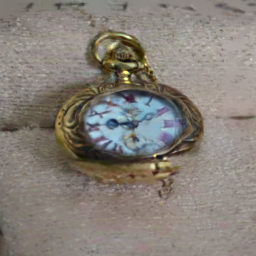} &
        
        \adjincludegraphics[clip,height=0.11\linewidth,trim={0 0 0 0}]{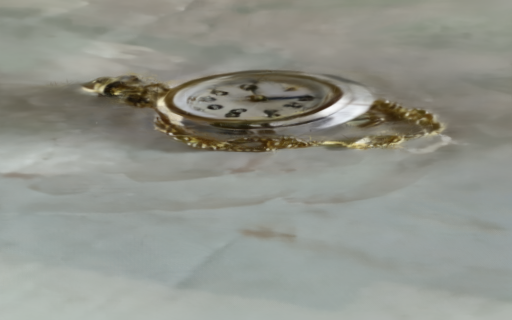} &
        \adjincludegraphics[clip,height=0.11\linewidth,trim={0 0 0 0}]{
        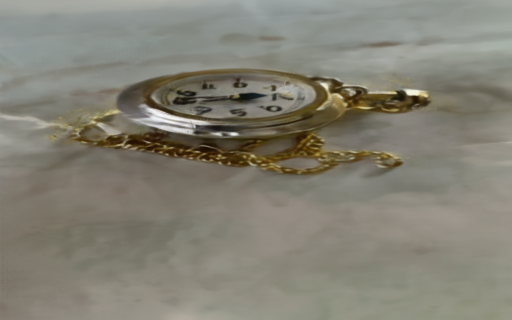} \\  [-2.6pt]
        \midrule
        \multirow{2}{*}{%
            \parbox{0.1\linewidth}{%
                \vspace{-18pt}%
                \centering
                \footnotesize
                \textit{A wall with two movie posters and a lighted sign that reads "4".}%
                    }%
        } 
        &
        \adjincludegraphics[clip,width=0.11\linewidth,trim={0 0 0 0}]{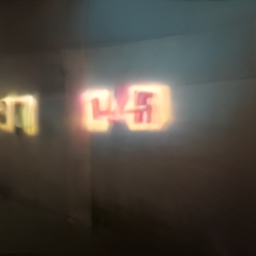} &
        \adjincludegraphics[clip,width=0.11\linewidth,trim={0 0 0 0}]{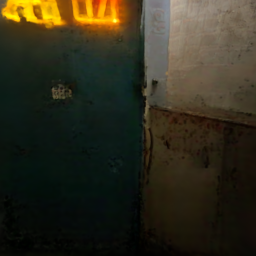} &
          \adjincludegraphics[clip,width=0.11\linewidth,trim={0 0 0 0}]{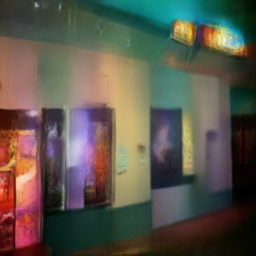} &
        \adjincludegraphics[clip,width=0.11\linewidth,trim={0 0 0 0}]{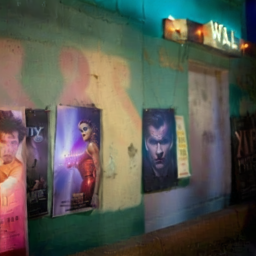} &
        \adjincludegraphics[clip,height=0.11\linewidth,trim={0 0 0 0}]{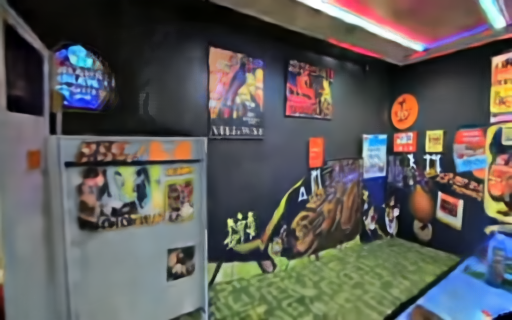} &
        \adjincludegraphics[clip,height=0.11\linewidth,trim={0 0 0 0}]{
        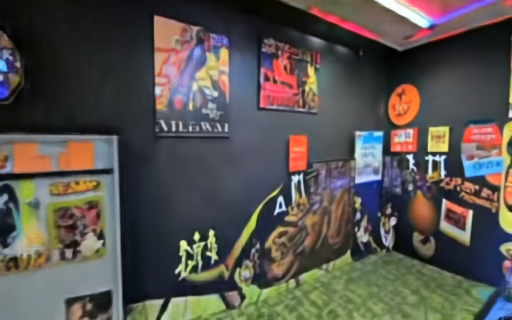
        } \\ [-2.6pt]

        & 
        \adjincludegraphics[clip,width=0.11\linewidth,trim={0 0 0 0}]{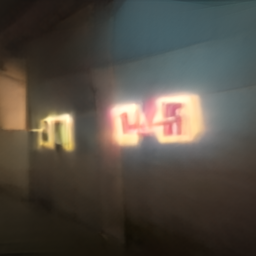} &
        \adjincludegraphics[clip,width=0.11\linewidth,trim={0 0 0 0}]{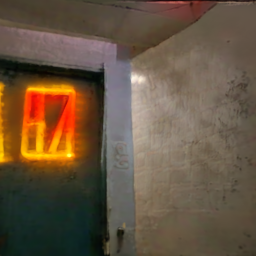} &
        
          \adjincludegraphics[clip,width=0.11\linewidth,trim={0 0 0 0}]{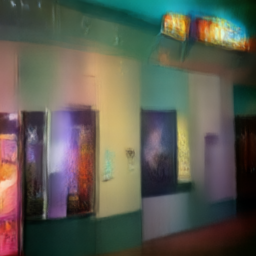} &
        \adjincludegraphics[clip,width=0.11\linewidth,trim={0 0 0 0}]{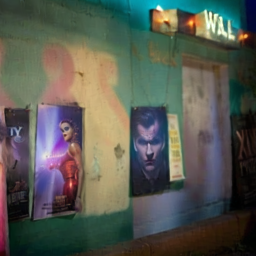} &
        \adjincludegraphics[clip,height=0.11\linewidth,trim={0 0 0 0}]{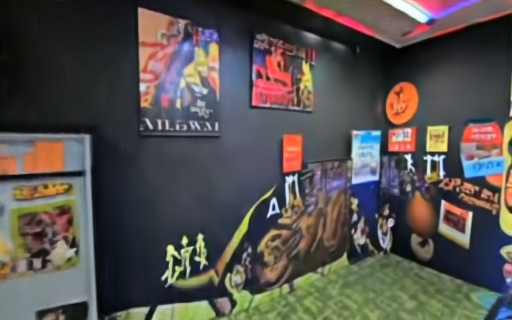} &
        \adjincludegraphics[clip,height=0.11\linewidth,trim={0 0 0 0}]{
        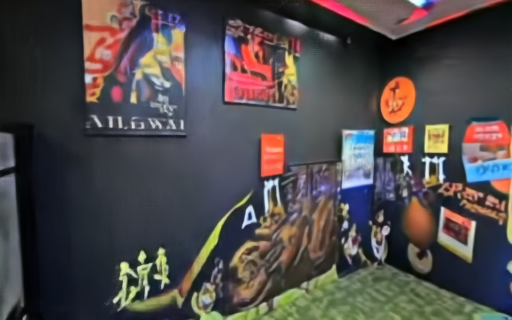} \\  [-2.6pt] \midrule
        \multirow{2}{*}{%
            \parbox{0.1\linewidth}{%
                \vspace{-20pt}%
                \centering
                \footnotesize
                \textit{.. a classical architectural structure with two prominent female statues on either side..}%
                    }%
        } 
        &
        \adjincludegraphics[clip,width=0.11\linewidth,trim={0 0 0 0}]{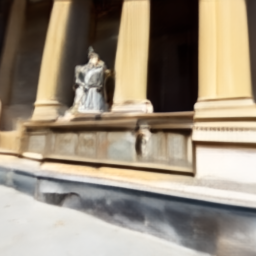} &
        \adjincludegraphics[clip,width=0.11\linewidth,trim={0 0 0 0}]{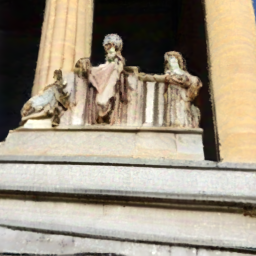} &
          \adjincludegraphics[clip,width=0.11\linewidth,trim={0 0 0 0}]{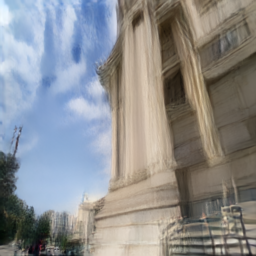} &
        \adjincludegraphics[clip,width=0.11\linewidth,trim={0 0 0 0}]{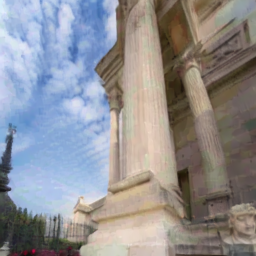} &
        \adjincludegraphics[clip,height=0.11\linewidth,trim={0 0 0 0}]{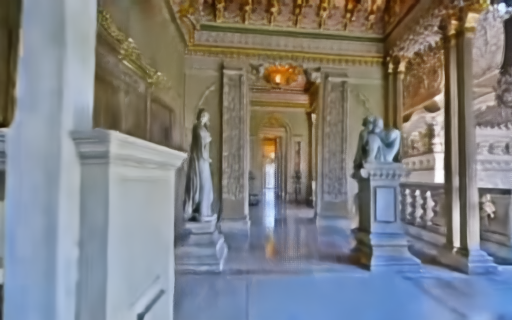} &
        \adjincludegraphics[clip,height=0.11\linewidth,trim={0 0 0 0}]{
        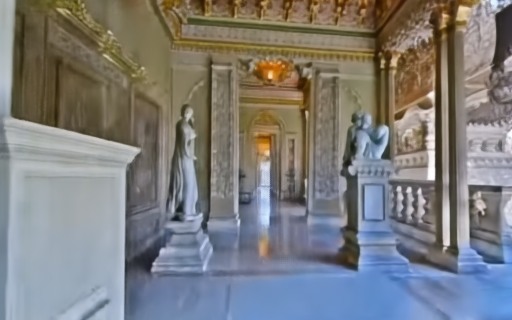
        } \\ [-2.6pt]

        & 
        \adjincludegraphics[clip,width=0.11\linewidth,trim={0 0 0 0}]{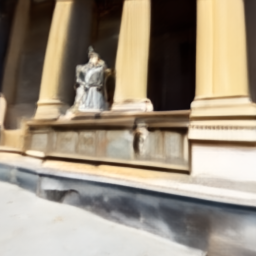} &
        \adjincludegraphics[clip,width=0.11\linewidth,trim={0 0 0 0}]{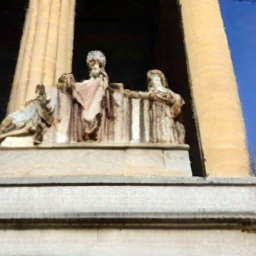} &
        
          \adjincludegraphics[clip,width=0.11\linewidth,trim={0 0 0 0}]{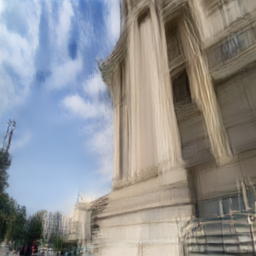} &
        \adjincludegraphics[clip,width=0.11\linewidth,trim={0 0 0 0}]{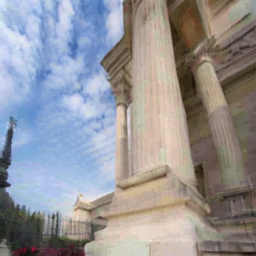} &
        \adjincludegraphics[clip,height=0.11\linewidth,trim={0 0 0 0}]{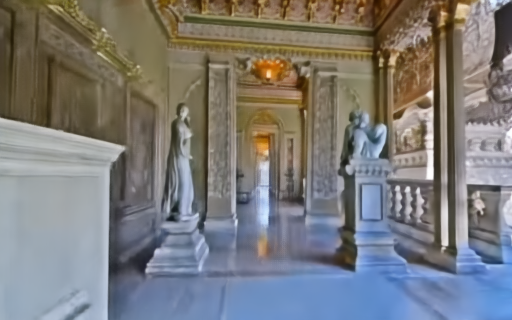} &
        \adjincludegraphics[clip,height=0.11\linewidth,trim={0 0 0 0}]{
        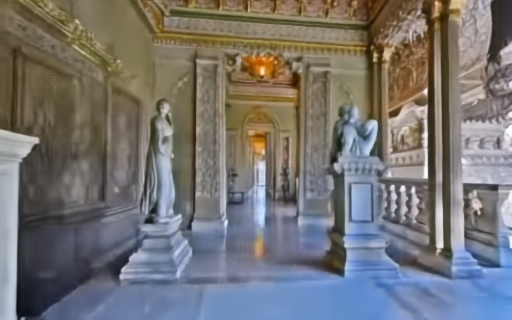} \\  [-2.6pt]

        & {\footnotesize Original} & {\footnotesize w/ SDS++} & {\footnotesize Original} & {\footnotesize w/ SDS++} & \multicolumn{2}{c}{\footnotesize Original}
        \\  [-1.0pt]
        \multicolumn{1}{c}{\small Text} & \multicolumn{2}{c}{\small Director3D~\cite{li2025director3d}} & \multicolumn{2}{c}{\small SplatFlow~\cite{go2024splatflow}} & \multicolumn{2}{c}{\small Ours}
    \end{tabular}
    \vspace{-0.2cm}
    \caption{
    \textbf{Additional qualitative comparison with Director3D~\cite{li2025director3d} and SplatFlow~\cite{go2024splatflow}}. Our VideoRFSplat generates more realistic scenes compared to baselines without relying on SDS++~\cite{li2025director3d}.
    }
    \vspace{-1mm}
    \label{fig:qual_scene_generation_additional}
\end{figure*}

We present additional qualitative comparison results of text-to-3DGS in Fig.~\ref{fig:qual_scene_generation_additional}.
Consistent with Fig. \textcolor{iccvblue}{5}, our VideoRFSplat can generate more realistic and detailed scenes than baselines without relying on SDS refinements.

\subsection{Additional Qualitative Results}
\begin{figure*}
    \centering
    \includegraphics[width=1.0\textwidth]{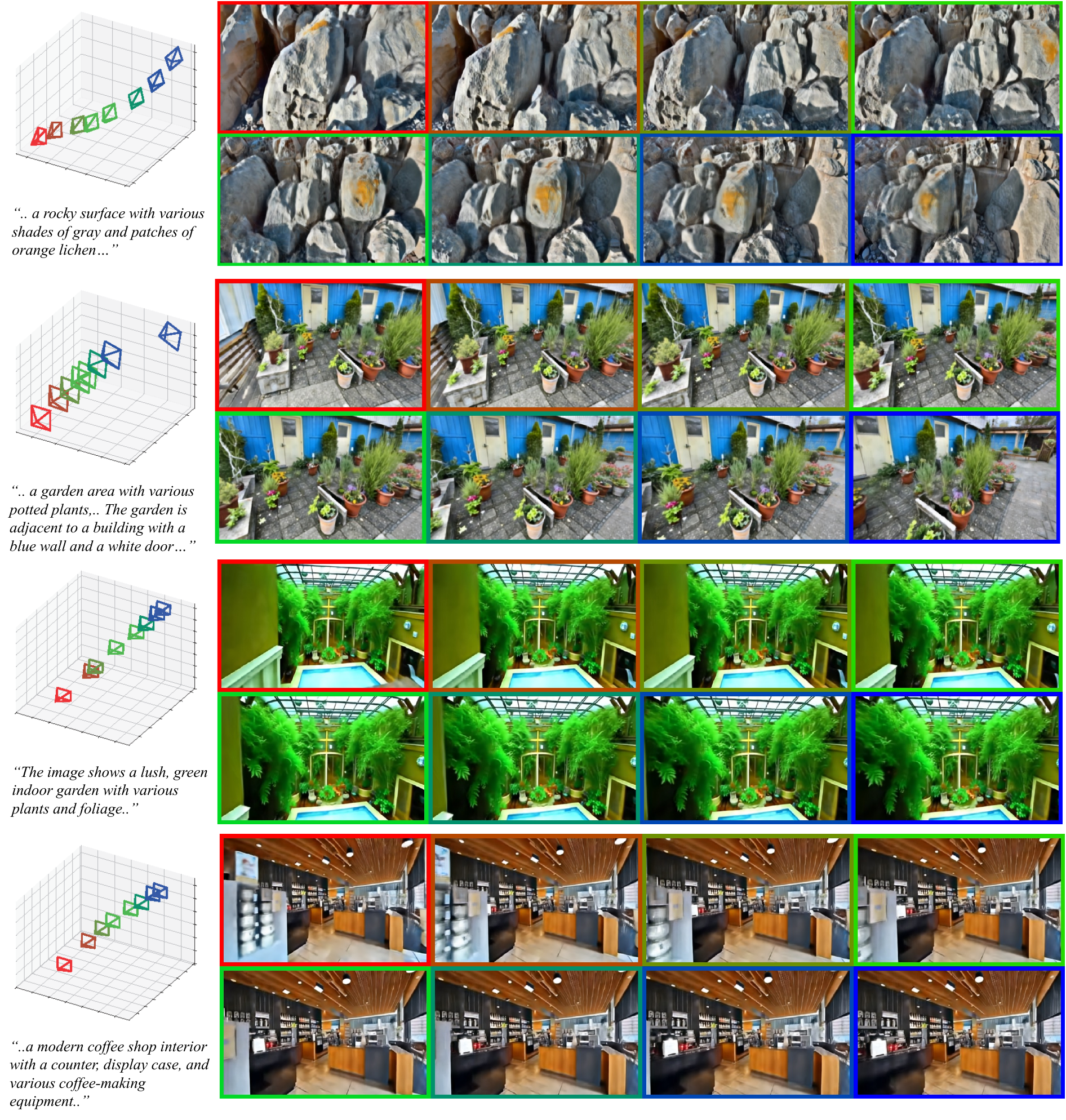}
    \caption{\textbf{Additional qualitative results.}We present eight rendered scenes along with their corresponding camera poses from text prompts, with image border colors indicating the respective cameras.}
    \label{fig:app_qual_1}
\end{figure*}

\begin{figure*}
    \centering
    \includegraphics[width=1.0\textwidth]{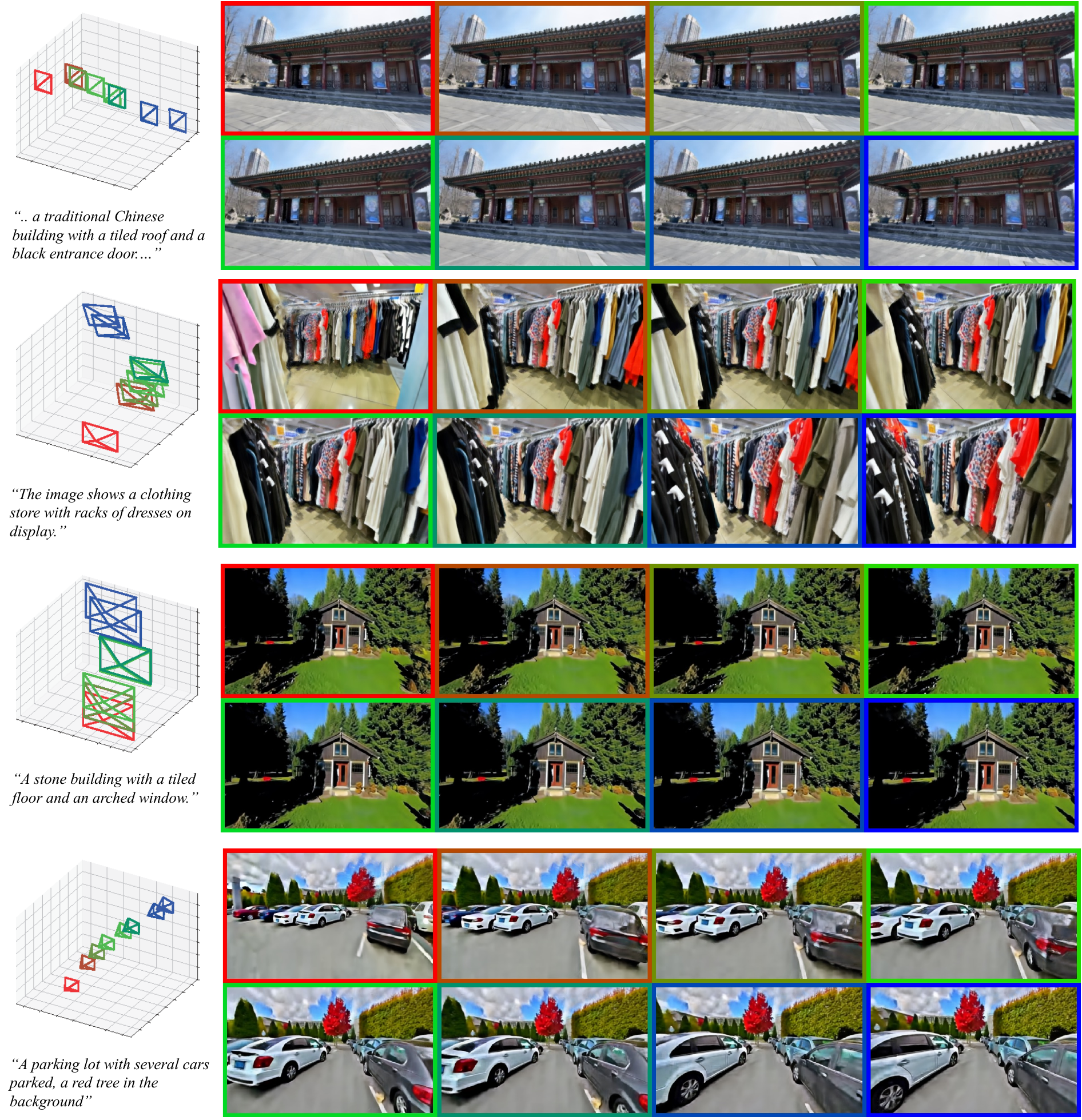}
    \caption{\textbf{Additional qualitative results.}We present eight rendered scenes along with their corresponding camera poses from text prompts, with image border colors indicating the respective cameras.}
    \label{fig:app_qual_2}
\end{figure*}

\begin{figure*}
    \centering
    \includegraphics[width=1.0\textwidth]{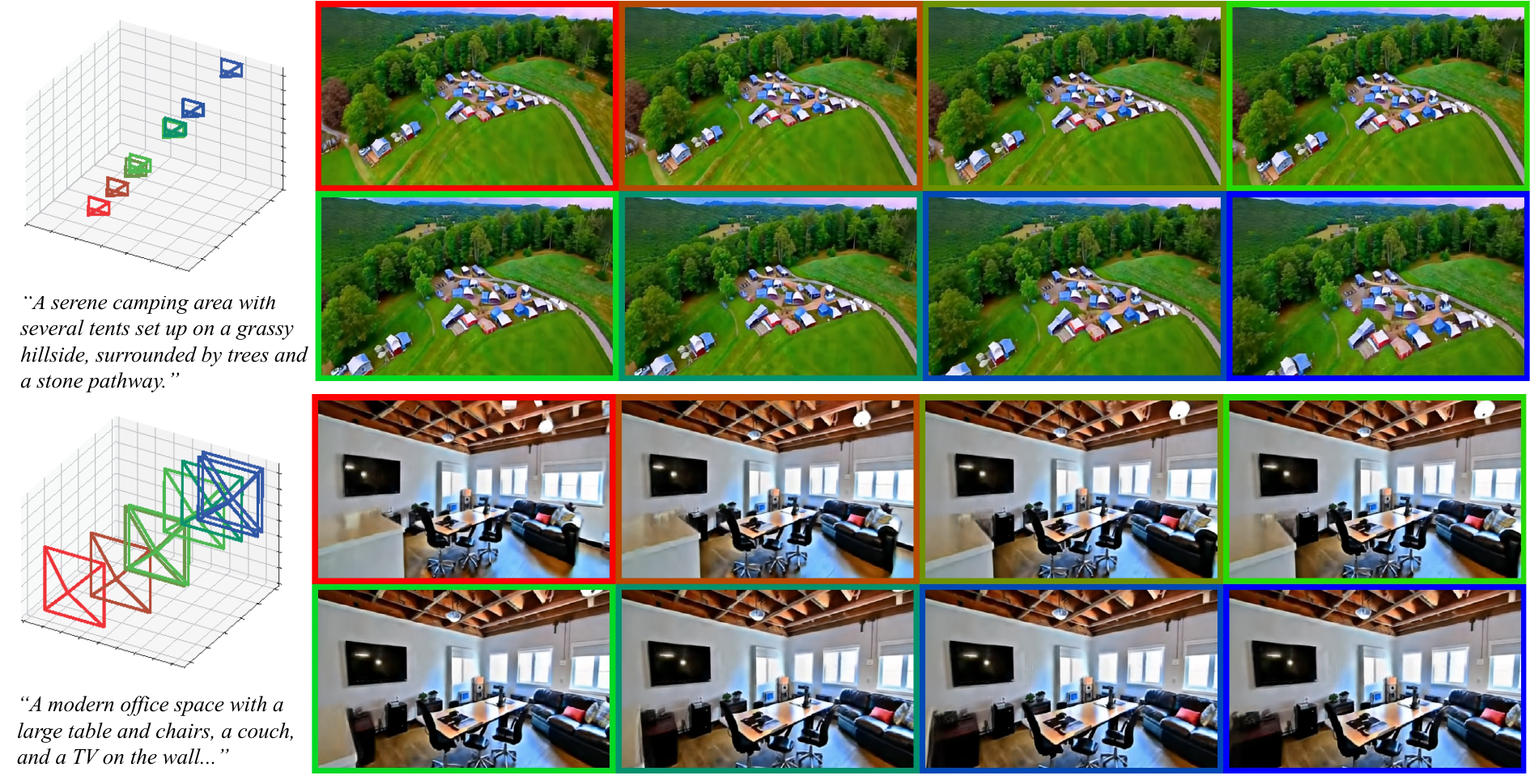}
    \caption{\textbf{Additional qualitative results.}We present eight rendered scenes along with their corresponding camera poses from text prompts, with image border colors indicating the respective cameras.}
    \label{fig:app_qual_3}
\end{figure*}

We illustrate additional qualitative results of VideoRFSplat in Fig.~\ref{fig:app_qual_1},~\ref{fig:app_qual_2}, and~\ref{fig:app_qual_3}.
As shown in the results, our VideoRFSplat can generate high quality 3DGS aligned with text prompts.

\subsection{Results on Camera Conditioned Generation}

\begin{figure*}
    \centering
    \includegraphics[width=1.0\textwidth]{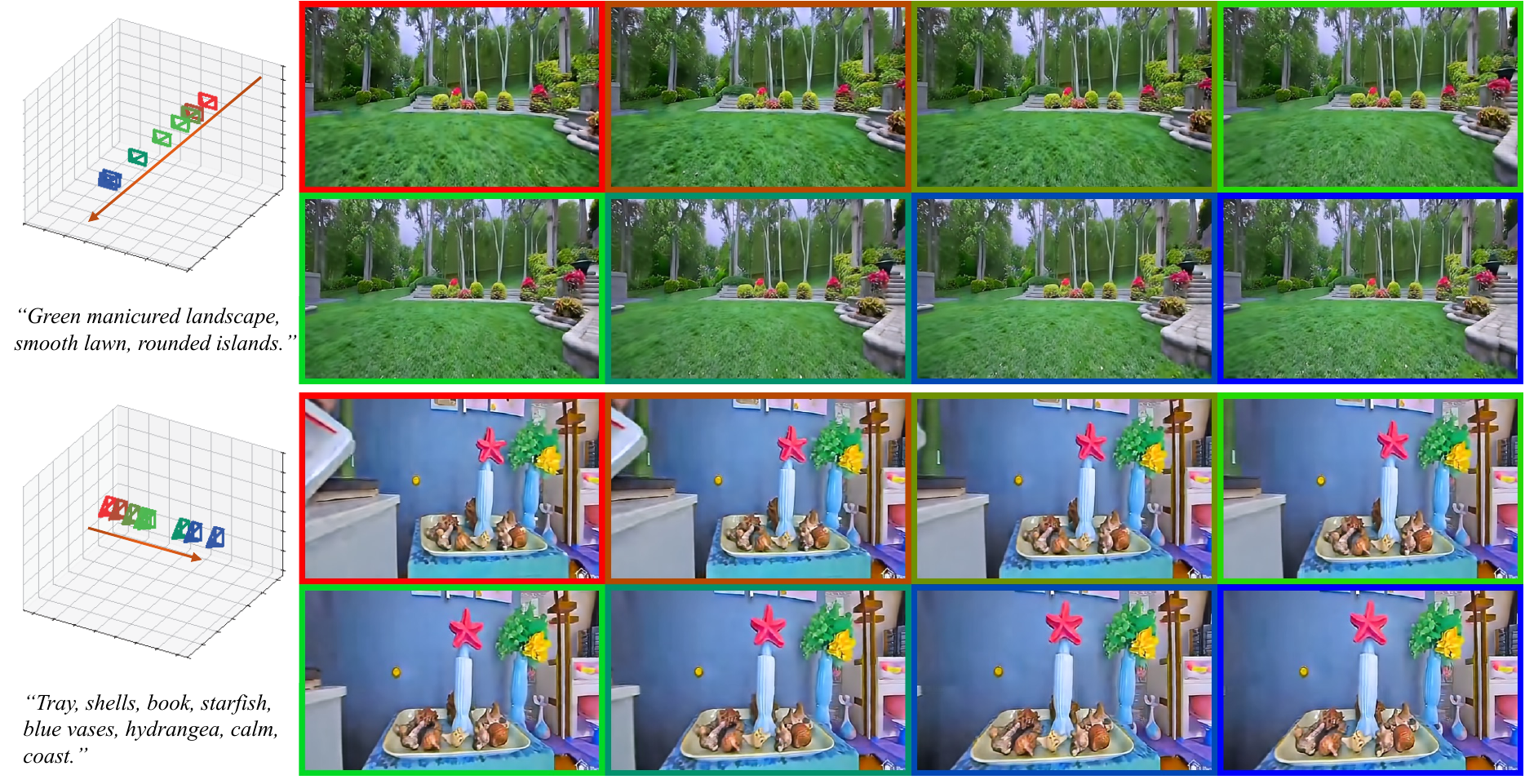}
    \caption{\textbf{Qualitative results on camera-conditioned generation.} We present generated multi-view images from given text and camera trajectory conditions. VideoRFSplat can perform camera-conditioned generation.}
    \label{fig:cam_ctrl}
\end{figure*}

Due to the limited space, we hereby supplement qualitative results on camera conditioned generation.
Figure~\ref{fig:cam_ctrl} shows that VideoRFSplat accurately generates images following camera trajectories while aligned with text prompts. 

\end{document}